\documentclass[runningheads]{llncs}

\usepackage{eccv}
\usepackage{eccvabbrv}
\usepackage{graphicx}
\usepackage{booktabs}
\usepackage[accsupp]{axessibility}
\usepackage{soul}
\usepackage{url}
\usepackage[utf8]{inputenc}
\usepackage{color}
\usepackage{multirow}
\usepackage{caption}
\usepackage{cuted}
\usepackage{amssymb}
\usepackage{bbding}
\usepackage{colortbl}
\usepackage{array}
\usepackage{arydshln}
\usepackage{siunitx} 
\usepackage{wrapfig}
\usepackage[colorlinks]{hyperref}
\usepackage{orcidlink}
\usepackage{authblk}

\definecolor{ForestGreen}{rgb}{0.13, 0.55, 0.13} 
\definecolor{WildStrawberry}{rgb}{1.0, 0.26, 0.64}
\definecolor{AirForceblue}{rgb}{0.36, 0.54, 0.66}
\definecolor{aliceblue}{rgb}{0.94, 0.97, 1.0}
\newcommand{\increase}[1]{
	\fontsize{6pt}{0.5em}\selectfont\color{ForestGreen}{$\uparrow$~\textbf{#1}}
}
\newcommand{\decrease}[1]{
	\fontsize{6pt}{0.5em}\selectfont\color{WildStrawberry}{$\downarrow$~\textbf{#1}}
}
\makeatletter
\def\@fnsymbol#1{\ensuremath{%
		\ifcase#1
		\or 
		\dagger
		\or 
		\ddagger
		\or 
		\mathsection
		\or 
		\mathparagraph
		\else 
		\@ctrerr  
		\fi}}   
\makeatother

\title{PathMMU: A Massive Multimodal Expert-Level Benchmark for Understanding and Reasoning in Pathology} 

\titlerunning{PathMMU}

\author{Yuxuan Sun\inst{1,2} \and
Hao Wu\inst{3} \and
Chenglu Zhu\inst{2} \and
Sunyi Zheng\inst{2} \and
Qizi Chen\inst{4} \and
Kai Zhang\inst{5} \and
Yunlong Zhang\inst{1,2} \and
Dan Wan\inst{6} \and
Xiaoxiao Lan\inst{1} \and
Mengyue Zheng\inst{2} \and
Jingxiong Li\inst{1,2} \and
Xinheng Lyu\inst{2} \and
Tao Lin\inst{2,}\protect\footnotemark[1] \and
Lin Yang\inst{2,}\thanks{Corresponding author.}
}

\authorrunning{Y.~Sun et al.}
\institute{Zhejiang University \and
Westlake University\and
Macau University of Science and Technology \and
Jiangnan University \and
The Ohio State University \and
Fujian University of Traditional Chinese Medicine \\
\email{\{sunyuxuan,lintao,yanglin\}@westlake.edu.cn}
\\
\url{https://pathmmu-benchmark.github.io/}}

\begin{document}
\maketitle

\begin{abstract}
    The emergence of large multimodal models has unlocked remarkable potential in AI, particularly in pathology. However, the lack of specialized, high-quality benchmark impeded their development and precise evaluation. To address this, we introduce PathMMU, the largest and highest-quality expert-validated pathology benchmark for Large Multimodal Models (LMMs). It comprises 33,428 multimodal multi-choice questions and 24,067 images from various sources, each accompanied by an explanation for the correct answer. The construction of PathMMU harnesses GPT-4V's advanced capabilities,  utilizing over 30,000 image-caption pairs to enrich captions and generate corresponding Q\&As in a cascading process. Significantly, to maximize PathMMU's authority, we invite seven pathologists to scrutinize each question under strict standards in PathMMU's validation and test sets, while simultaneously setting an expert-level performance benchmark for PathMMU. We conduct extensive evaluations, including zero-shot assessments of 14 open-sourced and 4 closed-sourced LMMs and their robustness to image corruption. We also fine-tune representative LMMs to assess their adaptability to PathMMU. The empirical findings indicate that advanced LMMs struggle with the challenging PathMMU benchmark, with the top-performing LMM, GPT-4V, achieving only a 49.8\% zero-shot performance, significantly lower than the 71.8\% demonstrated by human pathologists. After fine-tuning, significantly smaller open-sourced LMMs can outperform GPT-4V but still fall short of the expertise shown by pathologists. We hope that the PathMMU will offer valuable insights and foster the development of more specialized, next-generation LMMs for pathology.

\end{abstract}
    
\section{Introduction}
\label{sec:intro}

\begin{figure*}[ht]
\centering
\includegraphics[width=\linewidth]{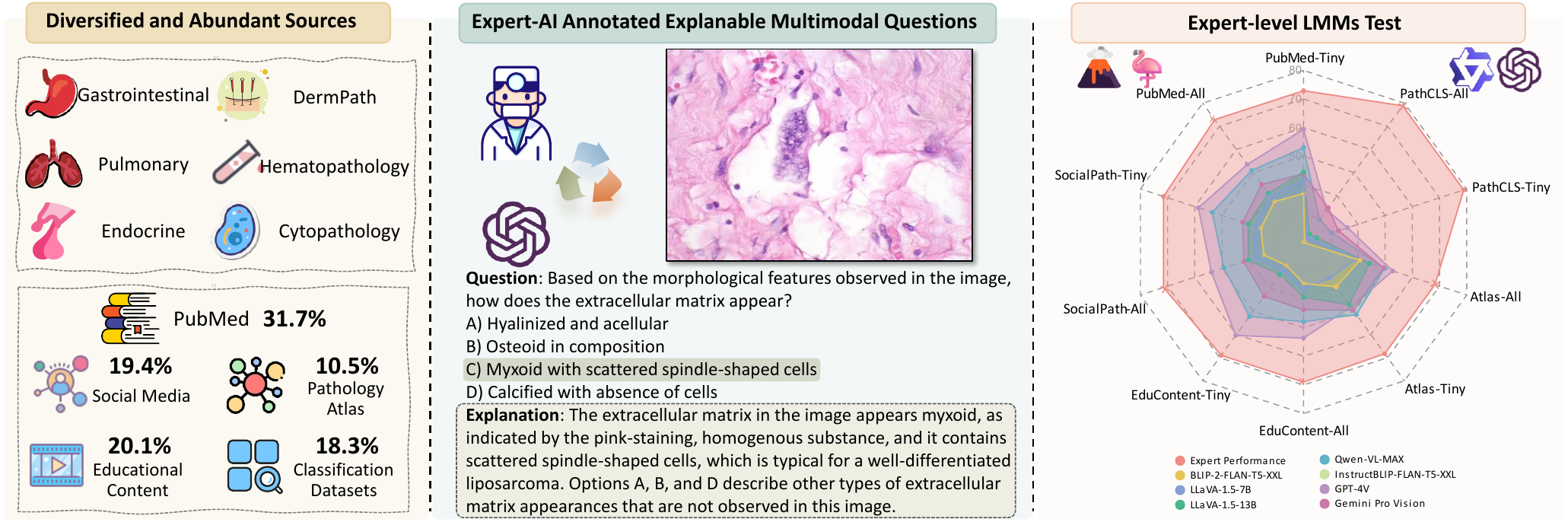}
\caption{An overview of the PathMMU benchmark: PathMMU is constructed using a diverse range of rich data sources. It comprises expert-level, multimodal, multi-choice questions in pathology, collaboratively crafted by AI and human pathology experts. Notably, even the most advanced LMMs substantially underperform when benchmarked against human experts on the PathMMU.}
\label{fig:overall}
\end{figure*}

Pathology is integral to modern medicine, serving as the foundation for diagnosing and understanding diseases. For example, liquid-based cytology plays a crucial role in early detection during cervical cancer screening. Concurrently, assessing biomarkers like HER2 in breast cancer and PD-L1 in various cancers in treatment informs the choice of targeted therapies and immunotherapies~\cite{kumar2014robbins}.

In recent years, the field of pathology has undergone a significant transformation, driven by advances in digital pathology and the rise of artificial intelligence (AI). This shift marks a departure from conventional microscope-based slide reading to AI-powered image analysis, greatly easing the workload of pathologists.
Traditional pathology models are tailored for specific tasks, such as cervical cytology and liver lesion classification,  leading to an abundance of task-specific models~\cite{zhu2022weakly,zhang2019pathologist}. In contrast, recent advancements in large multimodal models (LMMs) focus on offering general task-solving capabilities, thereby making universal pathological region recognition possible. This breakthrough not only represents a significant stride in the field, but also paves the way for more versatile and efficient diagnostic approaches in pathology~\cite{driess2023palm,li2023llava,sun2023pathasst}.

Given the demands of precision interpretation in the field of pathology, conducting comprehensive evaluations of LMMs' abilities in interpreting pathology images is essential. Yet, the field faces a notable scarcity of high-quality benchmark datasets. Currently, the representative large-scale dataset available is PathVQA, which contains over 30,000 visual question-answering samples with 4,998 images. Nevertheless, these samples are derived from a limited source of textbooks, with image captions being converted into questions through heuristic approaches, limiting the generation of diverse and logical Q\&As. Additionally, textbook captions often fail to accurately reflect the image contents. For instance, some captions might introduce supplementary details that are not evident in the images or partially cover the image content, thereby leading to unsolvable Q\&A pairs or questions that inadequately capture the essence of the images. Moreover, the absence of expert review and validation in the data curation process may introduce substantial noise into the dataset. These challenges pose significant obstacles to effectively validating LMMs' capabilities in pathology.

To address these challenges, we introduce PathMMU, an multimodal expert-level benchmark designed to evaluate LMMs in pathology image understanding and reasoning, comprising 33,428 Q\&As along with 24,067 pathology images. As depicted in \cref{fig:overall}, PathMMU draws from a diverse range of high-quality sources, including PubMed scientific documents, pathology atlas from textbooks and guidelines, Twitter posts by pathology experts, commonly used pathology classification datasets and educational teaching content from YouTube videos, covering multiple organ systems (\eg, gastrointestinal, pulmonary, endocrine, \etc) and multiple subjects (\eg, dermatopathology, hematopathology, cytopathology, \etc). In developing PathMMU, we adopt a meticulously designed cascading approach. Initially, we prompt GPT-4V with collected image-caption pairs to enhance the original captions, crafting descriptions that delve deeper into details and highlight crucial morphological features. Subsequently, we use these enhanced descriptions along with the images to prompt GPT-4V to generate professional multi-choice, multimodal pathology Q\&As, each accompanied by detailed explanations for their answers.  To ensure the questions are specifically designed for multimodal pathology understanding, we employ a collaboration of multiple large language models (LLMs) to eliminate questions that can be solved or guessed using text alone. Most importantly, we invite seven pathology experts to manually review approximately 12,000 Q\&As from the test and validation sets of PathMMU. The above carefully executed procedures, ensure the generation of professional, logical, and high-quality Q\&As. 

We evaluate the zero-shot performance of 18 advanced LMMs and their robustness on the PathMMU test sets. We also fine-tune two representative LMMs on the training set to assess their transfer learning capabilities. The key insights from this extensive analysis are: 
\textbf{(1) Cutting-edge LMMs struggle with the PathMMU}, with 15 out of the 18 achieving no more than 40\% accuracy. Even the most advanced model currently available, GPT-4V, only attains an accuracy of 49.8\%. This reveals a notable discrepancy of nearly 20\% compared to professional pathologists, indicating significant deficiencies of current LMMs in the specialized field of pathology and emphasizing the challenging nature of the PathMMU benchmark.  
\textbf{(2) While these LMMs show considerable robustness in robustness tests, the full extent of their robustness is still in question.} On the one hand, given their limited performance, the scope for further performance degradation is somewhat constrained. On the other hand, these models tend to capture only superficial image features, and sometimes may not even utilize image information for reasoning. As a result, even when image corruption obscures minor details, the impact on these models' performance might not be significantly pronounced.   
\textbf{(3) The text-only performance of GPT-4 Turbo surpasses the top-performing open-source LMMs on the PathMMU benchmark. We observe that GPT-4 Turbo sometimes takes shortcuts to guess answers.} This involves choosing the correct answer based either on the options most commonly encountered in real-world pathological scenarios, or by identifying the most distinctive option, rather than conducting an in-depth analysis of the pathology image.
\textbf{(4) Fine-tuning general-purpose LMMs with substantial amounts of data significantly boosts their ability to comprehend and reason with pathological images}, enabling them to easily surpass the zero-shot performance of GPT-4V. However, there still remains a discernible gap between their performance and that of human experts.

\section{Related Works}
\label{sec:related}

\textbf{Multimodal Models.} 
The emergence of powerful large language models such as BERT~\cite{devlin2018bert}, GPT-3~\cite{brown2020language}, T5~\cite{raffel2020exploring}, LLaMa~\cite{touvron2023llama},  ChatGPT~\cite{chatgpt}, GPT-4~\cite{gpt4}, and Vicuna~\cite{vicuna2023} has significantly advanced the field of natural language processing (NLP). These models demonstrate exceptional general capabilities, motivating researchers to explore the integration of LLMs with vision models to develop versatile multimodal models. This integration is primarily achieved through the pretraining approach, leading to the creation of groundbreaking LMMs. Representative models of this approach include CLIP~\cite{radford2021learning}, Flamingo~\cite{alayrac2022flamingo}, BLIP-2~\cite{li2023blip}, and Fuyu~\cite{fuyu-8b}, all of which demonstrate remarkable multimodal understanding abilities. Furthermore, by adapting instruction-tuning techniques from the NLP domain, a variety of multimodal instruction-tuning datasets have been developed. This initiative aims to guide LMMs in generating outputs that are more controllable, practical, and adaptable to various tasks. Simultaneously, this approach facilitates the integration of LLMs into LMMs using lightweight fine-tuning, significantly reducing both costs and time. Key models that embody this advancement include GPT-4V~\cite{openai2023gpt4v}, Gemini Pro Vision~\cite{team2023gemini}, Qwen-VL~\cite{bai2023qwen}, LLaVA~\cite{liu2023visual}, MiniGPT-4~\cite{zhu2023minigpt}, BLIP-2~\cite{li2023blip} and InstructBLIP~\cite{instructblip}.

\noindent \textbf{LMM Benchmarks.} 
In the general domain, the rise of various large models has led to the creation of extensive benchmark datasets, designed to evaluate the universal capabilities of these models across a range of tasks and domains. Notable examples include LAMM~\cite{yin2023lamm}, LVLM-eHub~\cite{xu2023lvlm}, SEED~\cite{li2023seed}, MMBench~\cite{liu2023mmbench}, MM-Vet~\cite{yu2023mm} and BenchLMM~\cite{cai2023benchlmm}, which have been used to assess the basic perception abilities of large models. More recently, MMMU~\cite{yue2023mmmu}, a massive and challenging dataset covering 30 university-level subjects, has been developed. It is specifically designed to evaluate the general multimodal understanding and reasoning capabilities of LMMs.  Furthermore, HaELM~\cite{wang2023evaluation} and HallusionBench~\cite{guan2023hallusionbench} are proposed to evaluate the degree of hallucination in LMMs.

In the medical field, datasets such as VQA-RAD~\cite{lau2018dataset} and VQA-Med~\cite{ben2021overview} provide a collection of Q\&A pairs based on radiology images, facilitating the evaluation of the capabilities of AI in radiological diagnostics. Additionally, PMC-VQA~\cite{zhang2023pmc} is a large-scale dataset within this domain. It generates a vast number of Q\&A pairs by prompting ChatGPT with text-only image captions from PubMed.
In the pathology domain, limited dataset construction efforts have been performed to date.  The representative large-scale dataset available, PathVQA~\cite{he2020pathvqa}, is constructed in a manner similar to PMC-VQA. It utilizes a heuristic approach to generate questions based on the text-only captions of pathology textbook images, yielding over 30,000 Q\&As and 4,998 images. More recently, Quilt-VQA \cite{seyfioglu2023quilt} is proposed, which is constructed by automatically extracting question and answer pairs from the narrations of professional teachers in YouTube videos. This approach guarantees the professionalism of the Q\&A pairs, although it comprises a relatively smaller dataset with 1,283 Q\&As and 985 images. Nonetheless, datasets reliant solely on image captions for question generation, face several notable limitations: \textbf{(1)} The creation of questions typically requires the consideration of both the image and its accompanying caption. Relying only on captions can introduce inaccuracies, making some questions unanswerable or solvable using only the text, bypassing the need for image analysis. \textbf{(2)} Since captions tend to be simplistic and may not capture the detailed information evident in images, generating an excessive number of Q\&As from the same image's caption can lead to either too simplistic or insufficient questions. \textbf{(3)} Automatically generated questions, whether through predefined rules or by prompting ChatGPT, are susceptible to containing numerous errors.

\begin{figure*}[t!]
	\centering
	\begin{minipage}{0.32\textwidth}
		\centering
		\includegraphics[width=1.0\linewidth]{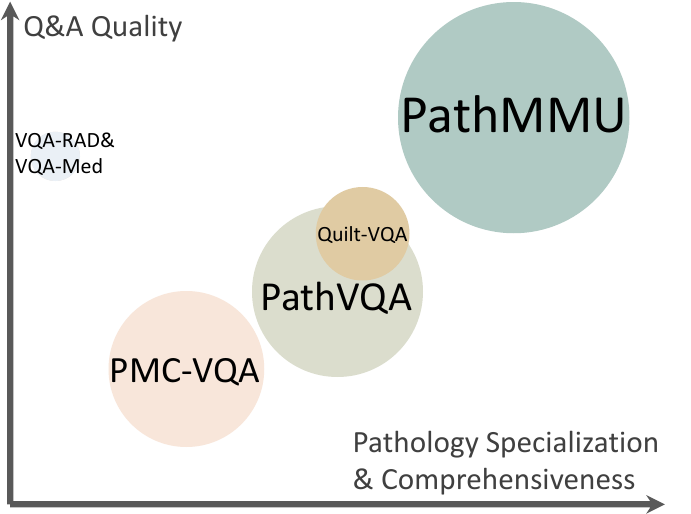}
	\end{minipage}\begin{minipage}{0.67\textwidth}
		\centering
		\captionsetup{type=table} 
		{\small
			\resizebox{\linewidth}{!}{
				\begin{tabular}{@{}ccccccccc@{}}
					\noalign{\hrule height 1.2pt}
				Dataset	& Domain    & Sources &   \# Q\&As / Images & \begin{tabular}[c]{@{}c@{}}Answer \\ Explanation?\end{tabular} & \begin{tabular}[c]{@{}c@{}}LLM \\  Filtered?\end{tabular} & \begin{tabular}[c]{@{}c@{}}Expert \\ Annotated? \end{tabular} \\ \midrule
					VQA-Med & Medical   & MedPix & 5000 / 5000 & \XSolidBrush & \XSolidBrush & \Checkmark   \\
					VQA-RAD & Medical   & MedPix & 3515 / 315  & \XSolidBrush & \XSolidBrush & \Checkmark   \\
					PMC-VQA & Medical   & PMC    & 227k / 149k & \XSolidBrush & \XSolidBrush & \XSolidBrush \\ \midrule
					PathVQA & Pathology & Book   & 32799 / 4998  & \XSolidBrush & \XSolidBrush & \XSolidBrush \\ 
                    \begin{tabular}[c]{@{}c@{}} Quilt-VQA \\Quilt-Red \end{tabular} & Pathology & Edu-Content & 1283 / 985  & \XSolidBrush & \XSolidBrush & \Checkmark \\ 
                    \addlinespace
					\rowcolor{aliceblue!60}
					PathMMU & Pathology & \begin{tabular}[c]{@{}c@{}} Social Media \\Edu-Content, PubMed \\ Atlas, CLS-Data \\ \end{tabular} & 33428  / 24067  &  \Checkmark & \Checkmark & \Checkmark  \\ \noalign{\hrule height 1.2pt}
			\end{tabular}}
		}
	\end{minipage}
	\caption{The comparison between PathMMU and existing benchmarks. The Q\&A pairs in PathMMU are sourced extensively and comprehensively, undergoing rigorous multi-tiered filtering. This includes the initial filtering by multiple  LLMs and the strict reviews by professional pathologists. Additionally, each question is accompanied by a detailed explanation. These attributes establish PathMMU as the most professionally curated, comprehensive, and highest-quality large-scale pathology dataset available.}
	\label{fig:comaprison}
\end{figure*}

Given the current situation of limited and low-quality benchmarks in the pathology domain, we develop PathMMU, a large-scale, high-quality, and expert-verified pathology benchmark, to fill this gap.
Our approach encompasses the generation of Q\&As derived from a diverse range of image-caption pairs, with subsequent rigorous expert review and filtering applied to both the images and the questions. Moreover, we provide detailed explanations for each answer as a reference to enhance the interpretability of answers, which is absent in previous works. 
As shown in \cref{fig:comaprison}, PathMMU distinguishes itself in the pathology domain through its professionalism and superior quality.

\section{The Proposed PathMMU Benchmark}
\subsection{Design Principle}
PathMMU is designed to provide the community with a specialized dataset for evaluating pathology LMMs. We adhere to five principal guidelines in its development: 
\textbf{(1) Comprehensive and Specialized Data}:
Our benchmark is gathered from a diverse range of authoritative materials, including respected scientific publications from PubMed, pathology atlas from textbooks and guidelines, educational videos, pathologist-shared images with explanations on Twitter, and widely recognized pathology classification datasets. Additionally, we involve seven human experts in manually reviewing the dataset to ensure it meets professional standards.
\textbf{(2) Effective and Valuable Questions}:
The questions in PathMMU are carefully formulated to necessitate answers based on detailed observation of pathology images, rather than being solvable through mere textual interpretation or guesswork. We ensure that these questions are answerable, effective, and aligned with the standards of professional pathology examinations.
\textbf{(3) Large-Scale}: We present the largest pathology dataset currently available. This scale enables researchers to thoroughly explore the full potential of LMMs in the field of pathology.
\textbf{(4) High-Quality Images}: We prioritize clarity in our images to ensure that every detail is discernible, as evidenced by our dataset's average resolution of approximately 900 $\times$ 700 pixels.
\textbf{(5) Explainable Answer Choices}: In line with the ongoing interest in understanding the decision-making logic of large models, our dataset includes reference explanations for each answer choice. This feature enhances the interpretability of model responses, contributing to the broader research on model explainability.

\subsection{The Construction of PathMMU}
\label{sec:pathmmu_construction}
\begin{figure*}[t!]
	\centering
	\includegraphics[width=\linewidth]{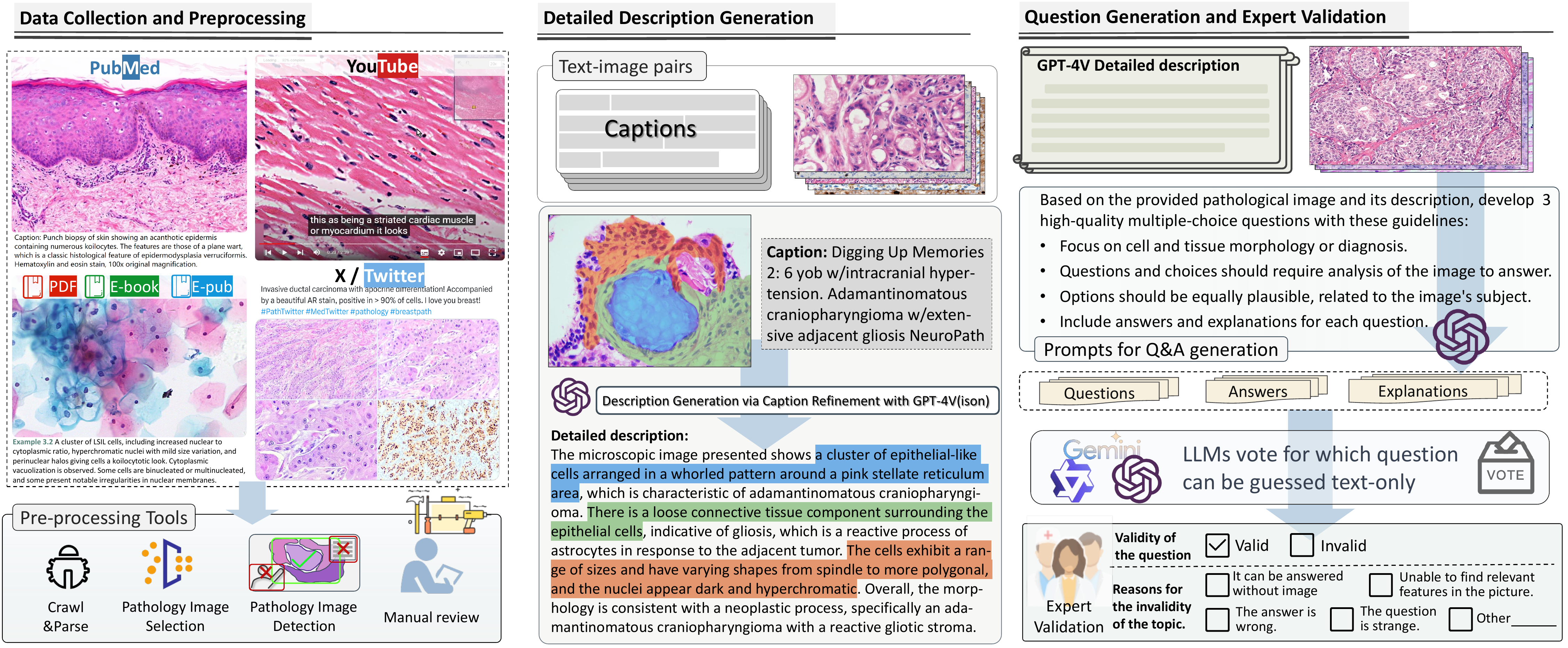}
	\caption{An illustrative overview of the three main processes in PathMMU Q\&A generation: data collection and preprocessing, detailed pathology image description generation, and question generation with LLMs filtering and expert validation.}
	\label{fig:qa_generation}
\end{figure*}
\label{sec:pathmmu construction}

To guarantee the quality of our benchmark, we meticulously develop a three-step data processing and generation protocol. The overall data collection and generation process is illustrated in \cref{fig:qa_generation}. 

\noindent \textbf{Step 1: Data Collection and Preprocessing.}
Our data collection draws from a wide variety of sources, which we integrate as subsets into the PathMMU framework. We name these subsets and detail their collection workflows as follows. 
\textbf{PubMed}: This subset consists of scientific pathology image-text pairs sourced from scientific documents. We gather these pairs from the open-access section of PubMed Central. Given the significant presence of non-pathological data within this resource, we meticulously annotate 20,000 samples as either pathological or non-pathological, and subsequently train a ConvNeXt~\cite{liu2022convnet} model to identify pathology data within the remaining dataset. 
\textbf{EduContent}: Derived from educational YouTube teaching videos, this subset is sourced from Quilt1M~\cite{ikezogwo2023quilt}. Given that YouTube videos often contain a wide range of visuals not related to pathology, such as computer desktops and unrelated imagery, the initial phase involved the precise annotation of pathological regions in 3,000 YouTube video images. This critical step is followed by the training of a YOLOv6-based detector~\cite{li2022yolov6} to automate the identification of pathological regions within remaining YouTube content.
\textbf{Atlas}: This subset is a compilation of authoritative pathology textbooks and guidelines, we transform them from PDFs to HTML to facilitate the extraction of image-caption pairs.  
\textbf{SocialPath}:  This subset aggregates image-text pairs from Twitter posts by pathology experts, using Twitter URLs provided by OpenPath~\cite{huang2023visual}. \textbf{PathCLS}: Encompassing widely recognized pathology classification datasets, including PatchCamelyon~\cite{veeling2018rotation}, CRC-100K~\cite{kather2018100}, SICAPv2~\cite{silva2020going}, BACH~\cite{aresta2019bach}, Osteo~\cite{arunachalam2019viable}, SkinCancer~\cite{kriegsmann2022deep}, MHIST~\cite{wei2021petri}, WSSSLUAD~\cite{han2022wsss4luad}, LC-Lung and LC-Colon~\cite{borkowski2019lung}. We reformulate these datasets into a Q\&A format and randomly sample from these datasets. We directly make the gathered samples from PathCLS a part of the PathMMU dataset without undergoing the following two steps Q\&A generation process. Finally, we manually review and filter irrelevant or unclear images to ensure the quality of collected data. As a result, approximately 30,000 high-quality image-text pairs are obtained, forming the foundation of the PathMMU.

\noindent \textbf{Step 2: Detailed Description Generation and Question-answer Pairs Generation.} 
When dealing with data sources from platforms like Twitter and YouTube, it's common to find a weak correlation between an image and its accompanying caption. For instance, in the SocialPath dataset, captions might only describe the aesthetic appeal of the image, which is irrelevant to pathological findings.  Similarly, in other data sources, captions tend to describe only a fraction of the features present in the image. To tackle this issue, we prompt GPT-4V to focus on describing the morphology of cells and tissues. It is crucial to note that merely prompting GPT-4V to output image descriptions might lead to incoherent content. Therefore, we provide GPT-4V with original captions for reference, significantly reducing the incidence of such occurrences and thus ensuring the generation of more precise and relevant descriptions.

\noindent \textbf{Step 3: Question Generation and Expert Validation.} 
Following the generation of image descriptions, GPT-4V is tasked with creating three questions per image, each accompanied by multiple-choice options, the correct answer, and detailed explanations for the answer. The creation of these explanations serves a dual purpose:  it not only offers valuable insights into the model reasoning process, but also facilitates the subsequent manual review phase.

It's crucial to note that despite the careful design of question generation prompts for GPT-4V, the text-only  GPT-4 is capable of correctly answering over 60\% of these questions through educated guesses, bypassing the need for visual cues. GPT-4 explains that its deductions are drawn from patterns such as one option being more common in typical pathological scenarios or exhibiting noticeable differences from other options. We delve deeper into this phenomenon in \cref{sec:analysis of llms guess answer} to offer a more thorough experimental analysis. To ensure that LMMs do not rely on educated guesses, thereby obscuring their true multimodal capabilities, we employ GPT3.5 Turbo, GPT-4 Turbo, Gemini Pro, and ERNIE-Bot-4 to perform educated guesses. Questions that at least three of these models correctly guess the answer are subsequently excluded.

We partitioned the remaining dataset into training, validation, and test sets. The training set comprises 16,312 images and 23,041 Q\&As, while the validation and test sets contain roughly 12,000 Q\&As and over 8,000 images, respectively. For the validation and test sets, we invite seven professional pathologists to conduct a thorough manual review. These pathologists initially attempt to answer the questions independently and then assess them based on the following criteria:
\textbf{(1)} Whether the question can be answered without an accompanying image. 
\textbf{(2)} Whether the answer can be inferred from the provided question and image.
\textbf{(3)} Whether the supplied answer is incorrect, if there is no correct answer, or if multiple correct answers are possible.
\textbf{(4)} Whether the generated question appears unusual or atypical compared to standard questions in pathology examinations.
Questions failing to meet these standards are deemed invalid and consequently removed from the PathMMU dataset.

Ultimately, after thorough expert review, we obtain 710 Q\&As accompanied by 510 images for the validation set, and 9,677 Q\&As with 7,213 images for the test set. Additionally, to establish a standard for expert performance, we extract a smaller subset from the test set, named `test-tiny', which includes 1,156 Q\&As. We invite two groups of professional pathologists to participate in this subset as an examination, and we average their performance to set a benchmark for expert performance in the PathMMU, serving as a comparison reference for LMMs.

\section{Experiments}
In this section, we carry out extensive experiments on PathMMU. Initially, we assess the zero-shot performance of cutting-edge LMMs. To verify whether LMMs effectively utilize the visual information of pathology images, we also evaluate the performance of text-only LLMs on PathMMU as a reference. Additionally, we apply common corruptions to the images in the test-tiny set to test the robustness of the LMMs against image corruptions. Furthermore, we select representative LMMs and evaluate their performance to explore their transfer-learning capabilities. Finally, we conduct experimental analysis on the issues of LLMs being able to guess answers identified during the data construction process.

\begin{table*}[!t]
	\caption{Overall results of models on the PathMMU \textbf{validation} and \textbf{test set}. Besides reporting the performance of LMMs, we add text-only LLM baselines that purely accept text as inputs. The best-performing LMM in each subset is \textbf{in-bold}, and the top-performing LLM is {\ul{underlined}}.}
	\centering
	\resizebox{\linewidth}{!}{
		\begin{tabular}{@{}lccccccccccccc@{}}
			\toprule
			\textbf{} & \textbf{\begin{tabular}[c]{@{}c@{}}Validation \\  Overall\end{tabular}} & \multicolumn{2}{c}{\textbf{Test Overall}} & \multicolumn{2}{c}{\textbf{PubMed}} & \multicolumn{2}{c}{\textbf{SocialPath}} & \multicolumn{2}{c}{\textbf{EduContent}} & \multicolumn{2}{c}{\textbf{Atlas}} &\multicolumn{2}{c}{\textbf{PathCLS}} \\ 
			&- & Tiny  & ALL  & Tiny  & ALL & Tiny  & All & Tiny  & All  & Tiny  & ALL  & Tiny  & ALL\\
			&(710) & (1156)  & (9677)  & (281)  & (3068) & (235)  & (1855) & (255)  & (1938)  & (208)  & (1007) & (177)  & (1809)\\\midrule
			
			\color{gray} Random Choice & \color{gray} 24.6 & \color{gray} 22.1 & \color{gray} 23.7 & \color{gray} 22.1 & \color{gray} 25.1 & \color{gray} 25.5 & \color{gray} 26.5 & \color{gray} 25.5 & \color{gray} 26.0 & \color{gray} 19.7 & \color{gray} 23.0  & \color{gray} 15.3 & \color{gray} 16.3\\ 
			\color{gray} Frequent Choice & \color{gray} 27.5 & \color{gray} 27.7 & \color{gray} 25.5 & \color{gray} 28.8 & \color{gray} 26.1 & \color{gray} 27.7 & \color{gray} 26.7 & \color{gray} 29.8 & \color{gray} 26.5 & \color{gray} 28.4 & \color{gray} 27.5 & \color{gray} 22.0 & \color{gray} 21.0  \\

			\rowcolor{aliceblue!60}  Expert performance & - &  71.8 &  - &  72.9 & -& 71.5 &  - &  69.0 &  - & 68.3 &  - & 78.9 &  - \\
			\midrule
			\multicolumn{12}{c}{\textbf{Large Multimodal Models (LMMs): Text + Image as Input}} \\ \midrule
			OpenFlamingo2-9B         & 26.2 & 24.1 & 24.2 & 24.9 & 25.5 & 25.5 & 25.3 & 19.6 & 24.4 & 28.8 & 25.5 & 22.0 & 20.4 \\
			Kosmos2                  & 26.9 & 26.1 & 24.9 & 27.8 & 25.8 & 27.2 & 24.3 & 25.1 & 25.9 & 23.1 & 24.2 & 27.1 & 22.8  \\
                LLaMA-Adapter2-7B        & 26.2 & 26.6 & 26.4 & 26.0 & 28.3 & 27.7 & 26.5 & 27.1 & 28.0 & 30.8 & 27.4 & 20.9 & 20.5  \\
                MiniGPT4-Vicuna-13B      & 27.2 & 25.5 & 27.7 & 28.8 & 30.1 & 24.3 & 27.2 & 25.5 & 29.3 & 27.9 & 28.6 & 19.2 & 21.8  \\			
                Otter                    & 26.1 & 29.3 & 28.1 & 34.9 & 30.2 & 26.0 & 28.4 & 30.6 & 30.0 & 29.3 & 28.2 & 23.2 & 21.3  \\
                Fuyu-8B                  & 27.5 & 30.1 & 29.2 & 35.9 & 30.5 & 28.9 & 30.7 & 30.6 & 29.8 & 27.9 & 30.6 & 24.3 & 23.9  \\
			CogVLM                   & 29.9 & 30.6 & 29.7 & 32.0 & 32.1 & 31.1 & 30.4 & 30.6 & 29.9 & 35.1 & 33.7 & 22.6 & 22.4  \\
                Qwen-VL-7B               & 29.4 & 32.1 & 31.5 & 34.9 & 32.9 & 33.6 & 35.8 & 34.1 & 33.6 & 35.6 & 33.8 & 18.6 & 21.7  \\			
                BLIP-2 FLAN-T5-XL        & 32.8 & 32.6 & 31.8 & 35.9 & 34.6 & 34.9 & 33.6 & 33.7 & 32.8 & 34.6 & 37.4 & 20.3 & 20.8  \\
			BLIP-2 FLAN-T5-XXL       & 33.4 & 33.3 & 33.5 & 37.0 & 37.4 & 35.7 & 34.6 & 30.2 & 34.5 & 39.4 & 40.7 & 19.8 & 20.6  \\
			InstructBLIP-FLAN-T5-XL  & 33.1 & 34.9 & 31.8 & 37.0 & 33.2 & 35.3 & 33.9 & 36.5 & 33.3 & 37.0 & 36.7 & 26.6 & 23.1  \\
			InstructBLIP-FLAN-T5-XXL & 32.7 & 34.3 & 33.9 & 39.1 & 37.2 & 33.6 & 34.3 & 34.5 & 36.0 & 38.5 & 39.3 & 22.6 & 22.7  \\
   			LLaVA-1.5-7B             & 36.6 & 34.9 & 35.4 & 41.6 & 39.9 & 37.9 & 38.1 & 32.5 & 36.5 & 35.6 & 39.2 & 23.2 & 21.9  \\
			LLaVA-1.5-13B            & 37.9 & 38.8 & 37.6 & 44.5 & 41.0 & 40.4 & 40.4 & 34.1 & 39.4 & 47.1 & 44.3 & 24.9 & 23.5  \\
			\addlinespace\hdashline\addlinespace
			Qwen-VL-PLUS       & 38.0 & 39.3 & 34.3 & 43.5 & 37.7 & 41.3 & 36.0 & 39.6 & 36.0 & 44.7 & 37.1 & 23.2 & 23.3  \\
                Qwen-VL-MAX        & 43.6 & 49.2 & 45.9 & 53.0 & 50.9 & 53.6 & 49.3 & 52.2 & 47.9  & \textbf{51.4} & 49.8 & 30.5 & 29.6 \\
			Gemini Pro Vision  & 41.9 & 42.8 & 42.7 & 43.8 & 44.9 & 42.4 & 42.0 & 43.5 & 43.7 & 49.5 & 49.4 & 32.8 & \textbf{34.7}   \\
			GPT-4V             & \textbf{49.3} & \textbf{53.9} & \textbf{49.8} & \textbf{59.4} & \textbf{53.5} & \textbf{58.7} & \textbf{53.9} & \textbf{60.4} & \textbf{53.6} & 48.1 & \textbf{52.8}  & \textbf{36.2} & 33.8\\
			\midrule
			\multicolumn{12}{c}{\textbf{Large Language Models (LLMs): Only Text as Input}} \\ \midrule
                ERNIE-Bot 4.0 & 31.8 & 34.3 & 30.9 & 37.0 & 31.2 & 33.6 & 32.9 & 40.0 & 34.5 & 36.1 & 37.4 & 20.9 & 20.6\\ 			
                Gemini Pro    & 31.0 & 31.6 & 31.0 &31.3 & 33.5 & 31.9 & 31.4 & 33.3 & 32.7 & 38.0 & 37.7 & 21.5 & 20.9\\
			Vicuna-v1.5-13B & 32.0 & 31.2 & 31.6 & 35.2 & 33.8 & 30.6 & 34.3 & 31.0 & 33.5 & 34.6 & 35.7 & 22.0 & 20.7 \\
                GPT-3.5 Turbo & 30.4 & 29.4 & 28.4 & 32.7 & 31.2 & 31.5 & 30.1 & 31.0 & 30.1 & 26.9 & 26.8 & 22.0 & 20.8\\ 
                GPT-4 Turbo & \underline{36.5} & \underline{41.8} & \underline{38.1} & \underline{48.8} & \underline{42.4} & \underline{43.4} & \underline{42.3} & \underline{47.1} & \underline{40.6} & \underline{41.3} & \underline{43.2} & \underline{22.3} & \underline{21.1}\\     
			\bottomrule
		\end{tabular}
	}
	\label{tab:overall_results}
\end{table*}

\subsection{Zero-shot Evaluation of LMMs and LLMs}
In this study, we evaluate the zero-shot capabilities of the latest and most advanced LMMs on the PathMMU. Specifically, we use the PathMMU validation set to conduct prompt engineering for these models, followed by testing their performance on the test set. The tested model include 14 open-source LMMs: OpenFlamingo~\cite{alayrac2022flamingo},  Kosmos2~\cite{peng2023kosmos}, LLaMA-Adapter2~\cite{gao2023llama}, MiniGPT-4~\cite{zhu2023minigpt}, Otter~\cite{li2023otter}, Fuyu~\cite{fuyu-8b}, CogVLM~\cite{wang2023cogvlm}, Qwen-VL~\cite{bai2023qwen}, BLIP-2~\cite{li2023blip}, InstructBlip~\cite{instructblip}, LLaVA-v1.5~\cite{liu2023visual}, as well as 4 closed-source LMMs: Qwen-VL-PLUS, Qwen-VL-MAX~\cite{bai2023qwen}, Gemini Pro Vision~\cite{team2023gemini} and GPT-4V~\cite{openai2023gpt4v}. For BLIP-2, InstructBLIP, and LLaVA-v1.5, we deploy various sizes of these models to examine the applicability of scaling rules in LMMs. Moreover, we extend our evaluation to pure text-based LLMs, including ERNIE-Bot 4.0~\cite{sun2021ernie}, Gemini Pro~\cite{team2023gemini}, Vicuna-v1.5-13B~\cite{vicuna2023}, GPT-3.5 Turbo~\cite{chatgpt}, and GPT-4 Turbo~\cite{gpt4}. This assessment aims to quantify their ability to infer correct answers through educated guesses without visual inputs, reflecting the extent to which LMMs integrate and leverage image information compared to their text-only counterparts. 
\begin{figure*}[t!]
\centering
\begin{minipage}{0.41\textwidth}
    \centering
    \includegraphics[width=1.0\linewidth]{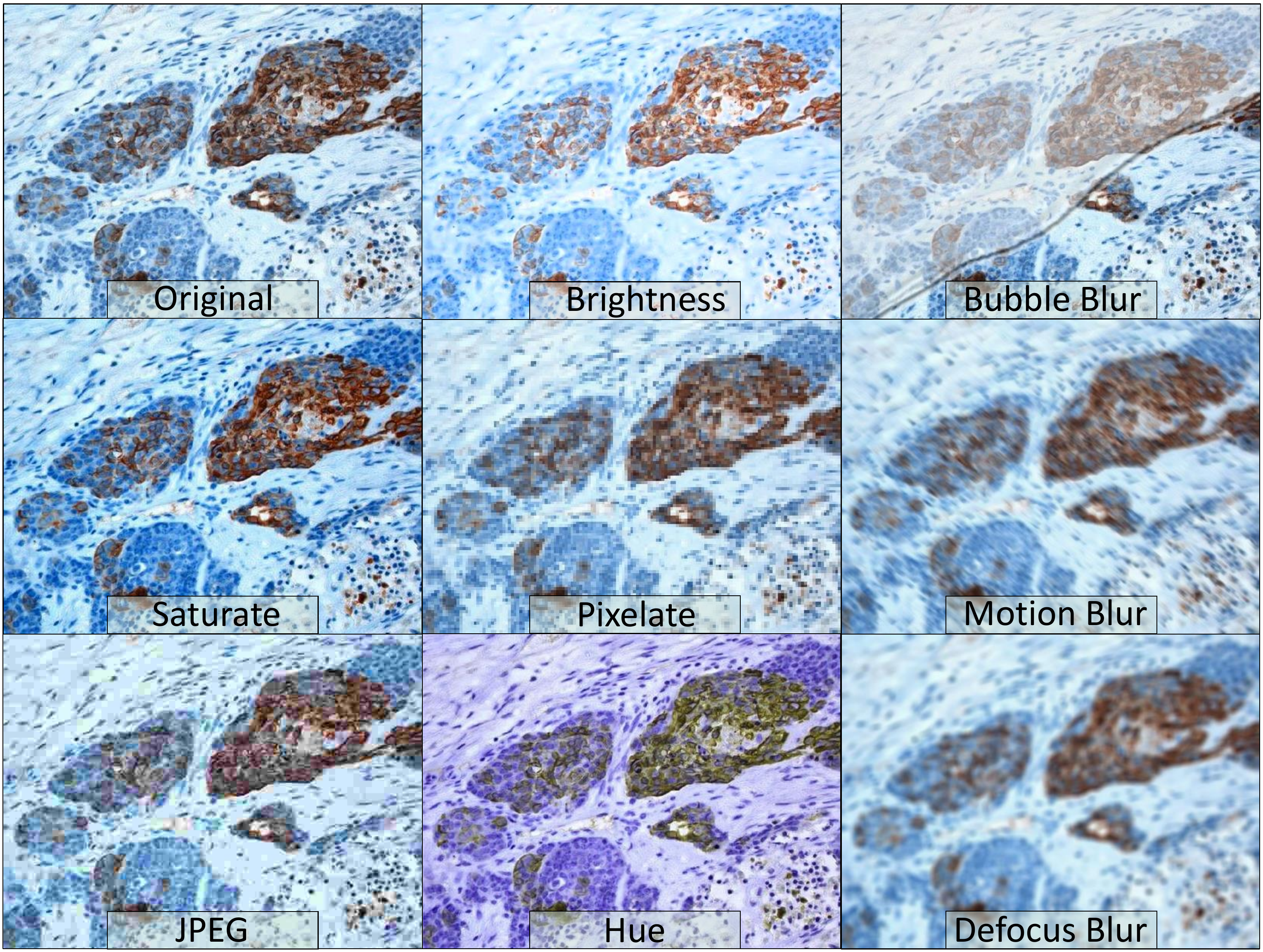}
\end{minipage}\begin{minipage}{0.58\textwidth}
    \centering
    \includegraphics[width=\linewidth]{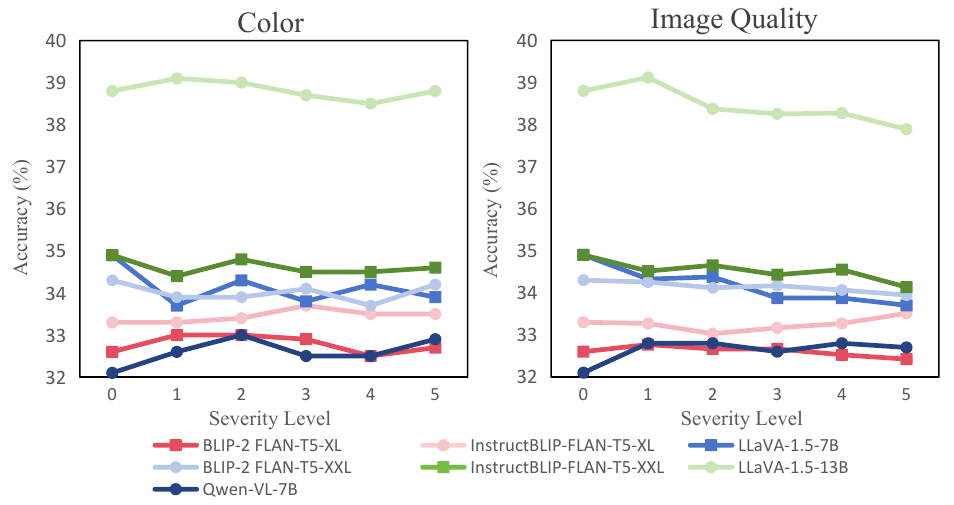}
\end{minipage}
\caption{Left: Illustration of corrupted pathology images. Right: LMM’s performance across various levels of color-related (brightness, hue, saturation) and image quality-related (pixelation, JPEG compression, bubble blur, motion blur, defocus blur) corruptions on the PathMMU test-tiny set, with level 0 representing the uncorrupted images.}
\label{fig:corruption&eval}
\end{figure*}

\begin{table*}[!t]
    \centering
    \caption{Evaluation of the model's robustness on the PathMMU test-tiny set, where \textcolor{ForestGreen}{green} indicates performance improvement and \textcolor{WildStrawberry}{red} signifies performance degradation compared to the model's performance with uncorrupted images.}
    \label{tab:overall_results of corruption}
    \resizebox{\linewidth}{!}{
        \begin{tabular}{lcccccccccccccccc}
            \toprule
           &\textbf{Brightness} &\textbf{Bubble} &\textbf{Defocus}  &\textbf{Hue} &\textbf{JPEG} &\textbf{Motion} &\textbf{Pixelate} &\textbf{Saturate} &\textbf{Overall} \\ 
            \midrule
            Qwen-VL-7B         & 32.7\increase{0.6} & 32.3\increase{0.2} & 32.1\decrease{0.0} & 32.4\decrease{0.3} & 32.3\increase{0.2} & 33.8\increase{1.7} & 33.1\increase{1.0} & 33.0\increase{0.9} & 32.7\increase{0.6} \\
            BLIP-2 FLAN-T5-XL  & 32.9\increase{0.3} & 32.4\decrease{0.2} & 32.8\increase{0.2} & 32.9\increase{0.3} & 32.4\decrease{0.2} & 32.8\increase{0.2} & 32.6\increase{0.0} & 32.7\increase{0.1} & 32.7\decrease{0.1} \\
            BLIP-2 FLAN-T5-XXL  & 33.1\decrease{0.2} & 33.2\decrease{0.1} & 33.5\increase{0.2} & 33.7\increase{0.4} & 33.5\increase{0.2} & 33.1\decrease{0.2} & 33.1\decrease{0.2} & 33.6\increase{0.3} & 33.3\decrease{0.0} \\
            InstructBLIP-FLAN-T5-XL  & 34.0\decrease{0.9} & 34.1\decrease{0.8} & 33.9\decrease{1.0} & 34.1\decrease{0.8} & 33.7\decrease{1.2} & 34.0\decrease{0.9} & 34.4\decrease{0.5} & 33.8\decrease{1.1} & 34.0\decrease{0.9} \\
            InstructBLIP-FLAN-T5-XXL & 34.0\decrease{0.3} & 33.8\decrease{0.5} & 34.2\decrease{0.1} & 34.1\decrease{0.2} & 34.3\decrease{0.2} & 34.3\increase{0.0} & 34.1\decrease{0.2} & 33.8\decrease{0.5} & 34.1\decrease{0.2} \\
            LLaVA-1.5-7B        & 34.8\decrease{0.1} & 35.3\increase{0.4} & 34.1\decrease{0.8} & 34.3\decrease{0.6} & 34.0\decrease{0.9} & 34.3\decrease{0.6} & 34.6\decrease{0.3} & 34.7\decrease{0.2} & 34.5\decrease{0.4} \\
            LLaVA-1.5-13B       & 38.9\increase{0.1} & 38.3\decrease{0.5} & 38.3\decrease{0.5} & 38.4\decrease{0.4} & 38.0\decrease{0.8} & 38.6\decrease{0.2} & 38.7\decrease{0.1} & 39.1\increase{0.3} & 38.5\decrease{0.3} \\
            \bottomrule
    \end{tabular}}%
\end{table*}

\noindent \textbf{Results and Discussion.} 
{\textit{\textbf{The advanced LMMs struggle with the PathMMU dataset}}. As shown in \cref{tab:overall_results}, among the 18 models tested, 15 show an accuracy below 40\%. Notably, the highest-performing open-source model, LLaVa-1.5-13B, and the top-performing closed-sourced model, GPT-4V, achieve only 37.6\% and 49.8\% accuracy, respectively. Those are significantly lower than human expert performance, which stands at 71.8\%, underscoring a substantial gap between current LMMs and professional pathologist standards. This gap indicates that the practical application of LMMs in real-world pathological scenarios remains significantly constrained.

\textit{\textbf{Closed-source LLMs achieve performance that is on par with or even surpasses that of the most advanced open-source LMMs.}} When considering text-only models, we prompt them to make educated guesses without giving the image. We observe that GPT-4 Turbo, Vicuna-v1.5-13B, Gemini-Pro, and ERNIE-Bot 4.0 outperform nearly half of the open-source LMMs. Notably, GPT-4 Turbo leads with a significant margin, achieving an accuracy rate of 38.1\%, exceeding the second-best LLM by 6.5\%, and even outperforming the best-performing open-source LMM, LLaVA-1.5-13B. The 38.1\% accuracy of GPT-4 Turbo, on the one hand, demonstrates that under our strict filtering criteria, even the best LLM, when provided with text alone, makes incorrect answers for the majority of questions. This underscores the effectiveness of our dataset in testing the capabilities of LMMs in pathological image analysis. On the other hand, it reaffirms the analysis in \cref{sec:pathmmu construction} that GPT-4 possesses strong logical reasoning abilities, enabling it to correctly guess a portion of the questions, significantly outperforming random selection. However, this brings up critical concerns in pathology diagnosis: the reliability of model outputs in clinical scenarios remains questionable, especially when they rely on educated guesses, particularly probabilistic ones, rather than comprehensive analysis. This emphasizes the urgent necessity to develop interpretable AI models and advocate for their cautious application in pathology diagnostics.

\textbf{\textit{The larger LMMs exhibit better performance.}} Specifically, LLaVa-1.5-13B exceeds LLaVa-1.5-7B model by 2.2\% in overall performance, while InstructBLIP-FLAN-T5-XXL and BLIP-2 FLAN-T5 XXL surpass their smaller versions by 2.1\% and 1.7\%, respectively. This suggests that larger models typically possess stronger multimodal capabilities in the pathology field.

\subsection{Robustness of LMMs to Image Corruption}
\label{sec:robustness}

In practical pathology, the interpretation of models significantly influences the subsequent medical decisions and treatment strategies. Therefore, models with strong robustness are crucial for clinical applications. The quality of pathological slides can be affected by various factors during staining, scanning, and storage, including JPEG compression, pixelation, blur (\eg, bubble blur, defocus blur, and motion blur), and color variations (\eg, brightness, saturation, and hue).

Drawing inspiration from the study on the robustness of encoder-based models to pathology image corruptions~\cite{zheng2024benchmarking,10095887,zhang2022benchmarking}, we incorporate these aforementioned types of corruptions into our analysis. To be specific, we simulate five levels of each corruption type on the PathMMU to explore the robustness of LMMs against these corruptions, as depicted in the left half of \cref{fig:corruption&eval}.

\begin{table*}[!t]
\caption{Results of LMMs on the PathMMU test set with the original images substituted by random Gaussian noise images.}
\resizebox{\linewidth}{!}{
    \centering
    \small
    \begin{tabular}{@{}lccccccccccccc@{}}
    \toprule
    \textbf{} &  \multicolumn{2}{c}{\textbf{Overall}} & \multicolumn{2}{c}{\textbf{PubMed}} & \multicolumn{2}{c}{\textbf{SocialPath}} & \multicolumn{2}{c}{\textbf{EduContent}} & \multicolumn{2}{c}{\textbf{Atlas}} &\multicolumn{2}{c}{\textbf{PathCLS}}\\ 
    & Tiny  & ALL  & Tiny  & ALL & Tiny  & All & Tiny  & All  & Tiny  & ALL  & Tiny  & ALL\\
    \midrule
    Qwen-VL-7B & 28.8\decrease{3.3} & 28.3\decrease{3.2} & 39.5\increase{4.6} & 30.4\decrease{2.5} & 24.7\decrease{8.9} & 30.5\decrease{5.3} & 27.8\decrease{6.3} & 28.9\decrease{4.7} & 27.4\decrease{8.2} & 30.1\decrease{3.7} & 20.3\increase{1.7} & 20.8\decrease{0.9} \\
    BLIP-FLAN-T5-XL         & 30.9\decrease{1.7} & 30.8\decrease{1.0} & 30.6\decrease{5.3} & 33.1\decrease{1.5} & 34.5\decrease{0.4} & 33.4\decrease{0.2} & 32.2\decrease{1.5} & 31.9\decrease{0.9} & 34.1\decrease{0.5} & 35.1\decrease{2.3} & 20.9\increase{0.6} & 20.8\decrease{0.0} \\
    BLIP-FLAN-T5-XXL        & 32.4\decrease{0.9} & 31.6\decrease{1.9} & 34.5\decrease{2.5} & 34.6\decrease{2.8} & 34.0\decrease{1.7} & 32.5\decrease{2.1} & 30.6\increase{0.4} & 32.8\decrease{1.7} & 38.5\decrease{0.9} & 38.1\decrease{2.6} & 22.2\increase{2.4} & 20.8\increase{0.2}  \\
    InstructBLIP-FLAN-T5-XL & 31.1\decrease{3.8} & 29.7\decrease{2.1} & 32.4\decrease{4.6} & 30.8\decrease{2.4} & 32.8\decrease{2.5} & 32.0\decrease{1.9} & 33.7\decrease{2.8} & 32.4\decrease{0.9} & 32.7\decrease{4.3} & 30.4\decrease{6.3} & 21.5\decrease{5.1} & 22.4\decrease{0.7} \\
    InstructBLIP-FLAN-T5-XXL& 30.4\decrease{3.9} & 30.8\decrease{3.1} & 32.0\decrease{7.1} & 33.5\decrease{3.7} & 28.9\decrease{4.7} & 31.9\decrease{2.4} & 32.2\decrease{2.3} & 31.7\decrease{4.3} & 35.1\decrease{3.4} & 35.2\decrease{4.1} & 21.5\decrease{1.1} & 21.7\decrease{1.0} \\
    LLaVA-1.5-7B            & 30.2\decrease{4.7} & 30.9\decrease{4.5} & 37.0\decrease{4.6} & 33.9\decrease{6.0} & 31.5\decrease{6.4} & 32.5\decrease{5.6} & 28.2\decrease{4.3} & 32.7\decrease{3.8} & 29.8\decrease{5.8} & 33.9\decrease{5.3} & 20.9\decrease{2.3} & 20.8\decrease{1.1} \\
    LLaVA-1.5-13B           & 32.7\decrease{6.1} & 33.2\decrease{4.4} & 38.4\decrease{6.1} & 36.0\decrease{5.0} & 35.3\decrease{5.1} & 36.3\decrease{4.1} & 29.0\decrease{5.1} & 34.1\decrease{5.3} & 35.6\decrease{11.5} & 38.8\decrease{5.5} & 22.0\decrease{2.9} & 21.0\decrease{2.5} \\

    \bottomrule
\end{tabular}}%
\label{tab:random image replace}
\end{table*}

\noindent \textbf{Results and Discussion.}  
\textit{\textbf{LMMs demonstrate notable robustness to various types and levels of image corruptions, yet their true robustness is questionable,}} as shown in \cref{tab:overall_results of corruption} and right half of \cref{fig:corruption&eval}. Notably, the Qwen-VL-7B even shows a 0.6\% overall performance increase under corruption compared to its baseline. It is more plausible to hypothesize that these corruptions mainly alter minute details of the pathology image that are difficult to discern with LMMs (such as chromatin morphology of the nucleus or vacuolization of the cytoplasm). This limitation stems from their pre-training in general domains, where they learn to recognize larger and more prominent features (\eg, humans, houses, cars, \etc) rather than the nuanced details that are vital in pathology. Furthermore, they may even resort to exploiting spurious correlations as shortcuts to answer the questions (\eg, deducing answers from textual patterns instead of focusing on essential image details.)

To substantiate our argument, we employ an extreme test of corruption by replacing the images with Gaussian noise images. As shown in \cref{tab:random image replace}, we discover that even when the images contain no relevant information, LMMs can still achieve results significantly better than random choice. The drop in performance when using these random images ranged only from 1.0\% to 4.5\%. This finding suggests that LMMs might rely on shortcuts to accomplish the multimodal task, such as depending exclusively on the textual information to make predictions. 
Additionally, it is intriguing to note that the performance drop across different sizes of the same model is remarkably similar. In other words, the benefit that images provide to models of various sizes appears to be consistent, suggesting that the improvement in performance for different-sized models may primarily stem from their LLM component, rather than the vision aspect.

\begin{figure*}[t!]
\centering
\begin{minipage}{0.45\textwidth}
    \centering
    \includegraphics[width=1.0\linewidth]{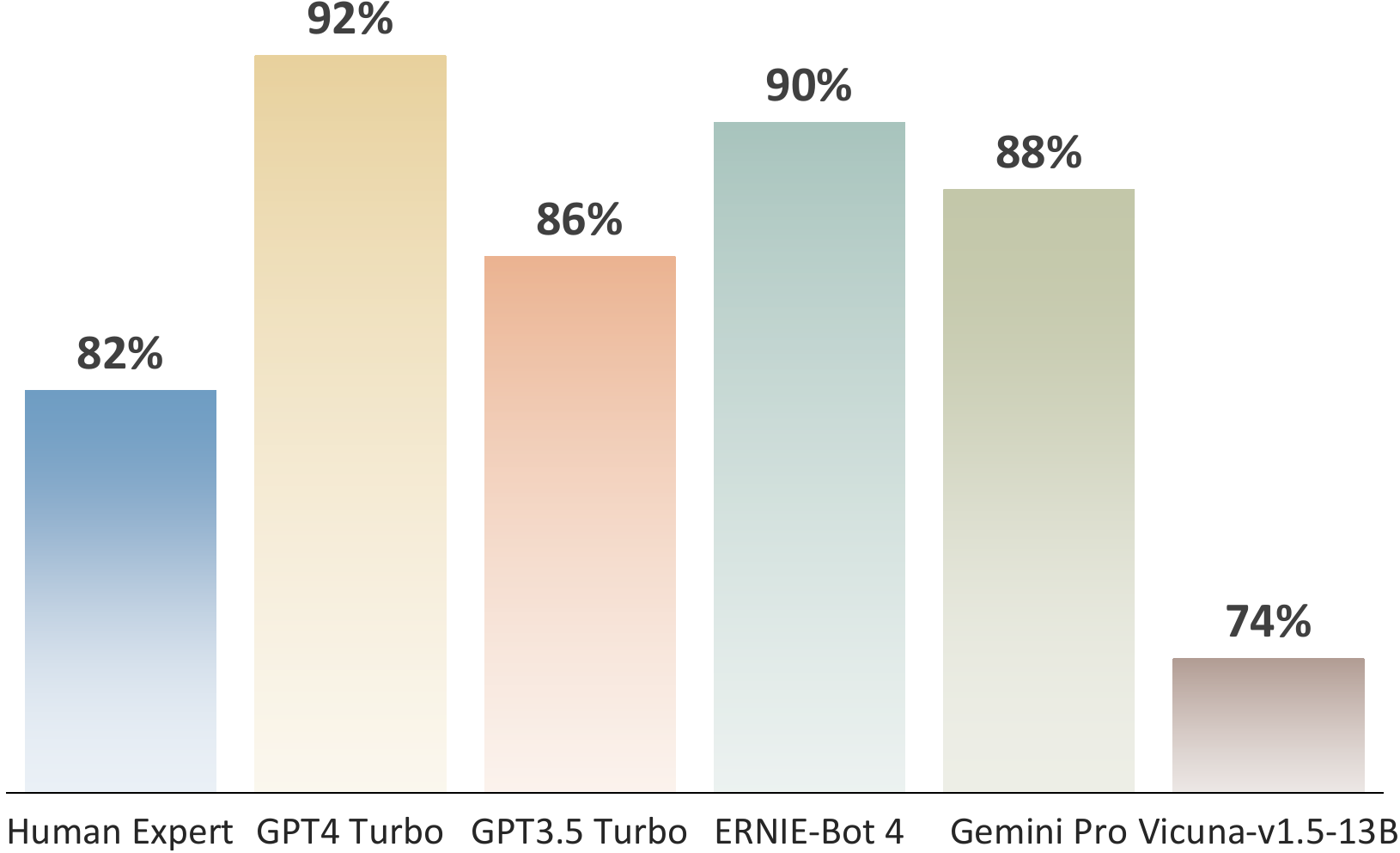}
\end{minipage}\hspace{0.1cm}\begin{minipage}{0.54\textwidth}
    \centering
    \captionsetup{type=table} 
    {\small
        \resizebox{\linewidth}{!}{%
            \begin{tabular}{@{}lccccc@{}}
                \toprule
                \multicolumn{1}{c}{} & \textbf{PubMed} & \textbf{SocialPath} & \textbf{EduContent} & \textbf{Atlas} & \textbf{Overall} \\ \midrule
                Random Choice                       & 25.8                & 24.3                & 25.5                & 25.8                & 25.3 \\
                \addlinespace\hdashline\addlinespace
                
                GPT-4 Turbo                         & 93.0 & 91.8 & 86.8 & 83.8 & 88.8 \\
                \quad  + Random Swap Question    & 44.3 & 45.5 & 50.0 & 51.5 & 47.8 \\
                GPT-3.5 Turbo                       & 84.3 & 86.5 & 79.5 & 85.3 & 83.9 \\
                \quad  + Random Swap Question    & 44.0 & 44.5 & 53.8 & 51.3 & 48.4 \\
                Gemini Pro                          & 85.5 & 90.1 & 76.5 & 71.5 & 84.1 \\
                \quad  + Random Swap Question    & 43.8 & 44.8 & 50.3 & 51.3 & 47.5 \\
                ERNIE-Bot 4.0                       & 94.5 & 92.3 & 89.3 & 82.0 & 89.5 \\
                \quad  + Random Swap Question    & 47.0 & 46.5 & 56.5 & 51.0 & 50.3 \\
                \addlinespace\hdashline\addlinespace
                BERT-large w/ NSP                      & 24.5  & 36.0 & 37.5 & 34.3  & 33.1 \\ 
                BiomedBERT-large w/ NSP                & 44.3 & 44.8 & 42.3 & 37.5 & 42.3 \\ \bottomrule
            \end{tabular}%
        }
    }
\end{minipage}
\caption{Left: The performance comparison between different LLMs and human experts on 100 filtered samples where the answer can be guessed through text-only. Right: Expand the sample size to 1600 to validate the source of LLM's ability to guess answers, which includes: (1) Randomly replacing the original questions with others from the dataset while keeping the options unchanged; and (2) utilizing the BERT series for answer selection, specifically through its Next Sentence Prediction (NSP), to assess whether an option is the sequential sentence following a question.}
\label{fig:llm_guess_analysis}
	
\end{figure*}
\subsection{Analysis of LLMs' Ability to Guess Answers}
\label{sec:analysis of llms guess answer}
In this section, we delve deeper into the phenomenon of LLM's ability to make educated guesses with more comprehensive experiments and analysis. 

We hypothesize several key reasons for the guessing behavior exhibited by LLMs: \textit{\textbf{(1) Frequency-based guessing:}} LLMs may guess based on the prevalence of certain options in real-world pathological instances. \textit{\textbf{(2) Based on the options that present a pattern of one supporting and three contradicting, or vice versa}}: For example, when a question identifies a tumour diagnosis, one option may match the pathological traits of a lesion, while the others suggest non-lesional characteristics. \textit{\textbf{(3) When only one option aligns with the question's object:}} For example, if a question asks which feature in the image supports the diagnosis of a Low-grade Squamous Intraepithelial Lesion (LSIL) cell, with only one option describes LSIL's pathological characteristics.

To support our hypothesis, we randomly select 100 samples filtered during the Q\&A generation process that can be correctly guessed by multiple LLMs. We then invite pathology experts to answer these questions, providing them with reference images. As shown in \cref{fig:llm_guess_analysis}, even with access to images, their performance is significantly lower than the closed-source LLMs, which reach approximately 90\%, while slightly better than the much smaller open-source Vicuna-v1.5-13B model. This suggests that the most advanced LLMs are superior to humans in guessing correct answers by identifying potential shortcuts within the questions. 

Furthermore, we expand the sample size to 1,600, with 400 from each source, to empirically analyze how LLMs guess answers. Specifically, we swap the questions among these samples while keeping the options unchanged, creating samples where the questions and options are entirely mismatched. As shown in the right half of \cref{fig:llm_guess_analysis}, we observe that LLMs still manage to guess about 50\% of these samples correctly, significantly higher than random choice accuracy. This finding suggests that the models tend to select the most common or prominent option as the answer, supporting our hypotheses (1) and (2).
To further explore hypothesis (3), we design experiments using BERT-large~\cite{devlin2018bert} and BiomedBERT-large~\cite{gu2021domain}, the latter being specifically pre-trained on biomedical data. 
Given that the BERT series includes next-sentence prediction (NSP) in their pretraining, which is inherently based on the similarity between two sentences, we apply this to predict the relationship between questions and options. We select the option with the highest probability of the next sentence to the question being the correct answer. Our findings reveal that both BERT and BiomedBERT substantially surpass random guessing, indicating that direct matching of questions and options is a viable method for models to guess answers, thereby supporting hypothesis (3). Moreover, BiomedBERT-large shows notable performance enhancement compared to BERT. This suggests that pretraining on biomedical data equips the model with a broader understanding of biomedical knowledge.

\subsection{Fine-tuning Results}
In order to explore LMMs' adaptability to the pathology domain, we select the representative LMMs, InstructBLIP-FLAN-T5-XL and InstructBLIP-FLAN-T5-XXL for fine-tuning experiments. Our experiments consist of two parts: (1) training the LMMs to directly generate answers, and (2) fine-tuning the LMMs to generate a reasoning process before delivering the final answer.

\noindent \textbf{Results and Discussion.}  
\textit{\textbf{All models exhibit significant improvements on the PathMMU test set after fine-tuning on its training set}}, as detailed in \cref{tab:overall_results of finetuning}. Notably, during the fine-tuning for direct answer generation, InstructBLIP-FLAN-T5-XL and InstructBLIP-FLAN-T5-XXL achieve significant improvements of 21.5\% and 21.3\%, respectively. This substantial improvement enables them to outperform the current leading model, GPT-4V. While still lagging behind expert performance, these results demonstrate a trend toward approaching expert-level proficiency. This also reflects the superiority of PathMMU in enhancing the LMMs' abilities for pathology image analysis.

Unexpectedly, generating explanations before answers during fine-tuning does not yield improvements. Instead, we observe slight performance decreases of 0.3\% and 2.3\% compared to direct answer generation, respectively.  We speculate that generating explanations is relatively more challenging, and incorporating it with answer generation might impede the models' capacity to generate correct answers. This finding raises an important question: how can we effectively leverage the interpretability information within PathMMU to enhance models’ training?

\begin{table*}[!t]
\caption{Results of LMMs on the PathMMU test set after the fine-tuning.  The `w/ A' and `w/ A\&E' denote the model is fine-tuned to output the answer directly or to output the answer with an explanation for its answer.}
\resizebox{\linewidth}{!}{
    \centering
    \small
    \begin{tabular}{@{}lccccccccccccc@{}}
        \toprule
        \textbf{} &  \multicolumn{2}{c}{\textbf{Overall}} & \multicolumn{2}{c}{\textbf{PubMed}} & \multicolumn{2}{c}{\textbf{SocialPath}} & \multicolumn{2}{c}{\textbf{EduContent}} & \multicolumn{2}{c}{\textbf{Atlas}} & \multicolumn{2}{c}{\textbf{PathCLS}} \\ 
        & Tiny  & ALL  & Tiny  & ALL & Tiny  & All & Tiny  & All  & Tiny  & ALL & Tiny  & ALL \\
        \midrule
        InstructBLIP-FLAN-T5-XL      & 34.9 & 31.8 & 37.0 & 33.2 & 35.3 & 33.9 & 36.5 & 33.3 & 37.0 & 36.7 & 26.6 & 23.1 \\
        \quad  + fine-tune w/ A      & 55.7\increase{20.8} & 53.3\increase{21.5} & 53.7\increase{16.7} & 50.4\increase{17.2} & 52.8\increase{17.5} & 52.3\increase{18.4} & 56.5\increase{20.0} & 52.0\increase{18.7} & 56.2\increase{19.2} & 53.1\increase{16.4} & 61.0\increase{34.4} & 60.8\increase{37.7}\\
        \quad  + fine-tune w/ A\&E   & 54.9\increase{20.0} & 53.0\increase{21.2} & 55.5\increase{18.5} & 51.4\increase{18.2} & 51.9\increase{16.6} & 49.7\increase{15.8} & 55.3\increase{18.8} & 51.5\increase{18.2} & 53.4\increase{16.4} & 51.6\increase{14.9} & 59.3\increase{32.7} & 61.4\increase{38.3}\\
        InstructBLIP-FLAN-T5-XXL     & 34.3 & 33.9 & 39.1 & 37.2 & 33.6 & 34.3 & 34.5 & 36.0 & 38.5 & 39.3 & 22.6 & 22.7\\
        \quad  + fine-tune w/ A      & 56.8\increase{22.5} & 55.2\increase{21.3} & 55.2\increase{16.1} & 51.5\increase{14.3} & 59.6\increase{26.0} & 55.2\increase{20.9} & 58.4\increase{23.9} & 54.1\increase{18.1} & 50.5\increase{12.0} & 53.7\increase{14.4} & 61.0\increase{38.4} & 63.7\increase{41.0}\\
        \quad  + fine-tune w/ A\&E   & 51.0\increase{16.7} & 52.9\increase{19.0} & 48.8\increase{9.7} & 50.8\increase{13.6} & 55.3\increase{21.7} & 51.2\increase{16.9} & 52.2\increase{17.7} & 51.9\increase{15.9} & 44.2\increase{5.7} & 50.1\increase{10.8} & 54.8\increase{32.2} & 60.5\increase{37.8}\\
        \bottomrule
\end{tabular}}%
\label{tab:overall_results of finetuning}
\end{table*}

\section{Conclusion}
In this study, we introduce PathMMU, the largest and highest-quality pathology benchmark to date, specifically crafted to evaluate the capabilities of LMMs in interpreting and reasoning with pathology images. The construction of PathMMU involves a meticulous data collection and curation process, supplemented by a strict manual review by seven professional pathologists to ensure its efficacy and professionalism. Moreover, we establish a human expert performance benchmark to quantify the gap between cutting-edge LMMs and human experts. Our experimental results reveal that advanced LMMs significantly lag behind on PathMMU, with these models demonstrating poor performance in observing details in pathology images and sometimes even neglecting visual information, highlighting a substantial gap in practical pathology application.  However, LMMs demonstrate notable performance improvements after fine-tuning on PathMMU, even surpassing GPT-4V. While they do not achieve human expert-level performance, these LMMs show promising potential for analyzing pathology images.

\textbf{Future Directions for Pathology LMMs:} Based on our experience with PathMMU, we identify several key areas where future large models in pathology need to focus: (1) Current multimodal models, which are primarily based on LLMs and fine-tuned in a lightweight manner, tend to over-rely on textual information while neglecting visual data. There is a significant need to explore training methodologies or model structures that better integrate visual and textual modalities. (2) There's a tendency for models to take shortcuts, solving problems in a ``lazy" manner. This necessitates the development of interpretable models to advance the creation of trustworthy models for real-world clinical applications. (3) 
Given that most current LMMs do not support multi-image inputs, PathMMU does not currently include a benchmark for processing multiple images. However, in practical scenarios, pathologists often analyze samples at various magnifications and from different perspectives, underscoring the importance of developing LMMs capable of handling multi-image inputs. Finally, we believe PathMMU will catalyze significant advancements in the development of next-generation LMMs in the pathology field.

%
%
\bibliographystyle{splncs04}
\bibliography{main}

\begin{thebibliography}{10}
\providecommand{\url}[1]{\texttt{#1}}
\providecommand{\urlprefix}{URL }
\providecommand{\doi}[1]{https://doi.org/#1}

\bibitem{alayrac2022flamingo}
Alayrac, J.B., Donahue, J., Luc, P., Miech, A., Barr, I., Hasson, Y., Lenc, K., Mensch, A., Millican, K., Reynolds, M., et~al.: Flamingo: a visual language model for few-shot learning. In: Advances in Neural Information Processing Systems (2022)

\bibitem{aresta2019bach}
Aresta, G., Ara{\'u}jo, T., Kwok, S., Chennamsetty, S.S., Safwan, M., Alex, V., Marami, B., Prastawa, M., Chan, M., Donovan, M., et~al.: Bach: Grand challenge on breast cancer histology images. Medical image analysis  \textbf{56},  122--139 (2019)

\bibitem{arunachalam2019viable}
Arunachalam, H.B., Mishra, R., Daescu, O., Cederberg, K., Rakheja, D., Sengupta, A., Leonard, D., Hallac, R., Leavey, P.: Viable and necrotic tumor assessment from whole slide images of osteosarcoma using machine-learning and deep-learning models. PloS one  \textbf{14}(4),  e0210706 (2019)

\bibitem{bai2023qwen}
Bai, J., Bai, S., Yang, S., Wang, S., Tan, S., Wang, P., Lin, J., Zhou, C., Zhou, J.: Qwen-vl: A frontier large vision-language model with versatile abilities. arXiv preprint arXiv:2308.12966  (2023)

\bibitem{fuyu-8b}
Bavishi, R., Elsen, E., Hawthorne, C., Nye, M., Odena, A., Somani, A., Ta\c{s}\i{}rlar, S.: Introducing our multimodal models (2023), \url{https://www.adept.ai/blog/fuyu-8b}

\bibitem{ben2021overview}
Ben~Abacha, A., Sarrouti, M., Demner-Fushman, D., Hasan, S.A., M{\"u}ller, H.: Overview of the vqa-med task at imageclef 2021: Visual question answering and generation in the medical domain. In: Proceedings of the CLEF 2021 Conference and Labs of the Evaluation Forum-working notes. 21-24 September 2021 (2021)

\bibitem{borkowski2019lung}
Borkowski, A.A., Bui, M.M., Thomas, L.B., Wilson, C.P., DeLand, L.A., Mastorides, S.M.: Lung and colon cancer histopathological image dataset (lc25000). arXiv preprint arXiv:1912.12142  (2019)

\bibitem{brown2020language}
Brown, T., Mann, B., Ryder, N., Subbiah, M., Kaplan, J.D., Dhariwal, P., Neelakantan, A., Shyam, P., Sastry, G., Askell, A., et~al.: Language models are few-shot learners. Advances in neural information processing systems  \textbf{33},  1877--1901 (2020)

\bibitem{cai2023benchlmm}
Cai, R., Song, Z., Guan, D., Chen, Z., Luo, X., Yi, C., Kot, A.: Benchlmm: Benchmarking cross-style visual capability of large multimodal models. arXiv preprint arXiv:2312.02896  (2023)

\bibitem{vicuna2023}
Chiang, W.L., Li, Z., Lin, Z., Sheng, Y., Wu, Z., Zhang, H., Zheng, L., Zhuang, S., Zhuang, Y., Gonzalez, J.E., Stoica, I., Xing, E.P.: Vicuna: An open-source chatbot impressing gpt-4 with 90\%* chatgpt quality (March 2023), \url{https://lmsys.org/blog/2023-03-30-vicuna/}

\bibitem{instructblip}
Dai, W., Li, J., Li, D., Tiong, A.M.H., Zhao, J., Wang, W., Li, B., Fung, P., Hoi, S.: Instructblip: Towards general-purpose vision-language models with instruction tuning (2023)

\bibitem{devlin2018bert}
Devlin, J., Chang, M.W., Lee, K., Toutanova, K.: Bert: Pre-training of deep bidirectional transformers for language understanding. In: Proceedings of the 2019 Conference of the North American Chapter of the Association for Computational Linguistics: Human Language Technologies, Volume 1 (Long and Short Papers). pp. 4171--4186 (2019)

\bibitem{driess2023palm}
Driess, D., Xia, F., Sajjadi, M.S., Lynch, C., Chowdhery, A., Ichter, B., Wahid, A., Tompson, J., Vuong, Q., Yu, T., et~al.: Palm-e: An embodied multimodal language model. arXiv preprint arXiv:2303.03378  (2023)

\bibitem{gao2023llama}
Gao, P., Han, J., Zhang, R., Lin, Z., Geng, S., Zhou, A., Zhang, W., Lu, P., He, C., Yue, X., et~al.: Llama-adapter v2: Parameter-efficient visual instruction model. arXiv preprint arXiv:2304.15010  (2023)

\bibitem{gu2021domain}
Gu, Y., Tinn, R., Cheng, H., Lucas, M., Usuyama, N., Liu, X., Naumann, T., Gao, J., Poon, H.: Domain-specific language model pretraining for biomedical natural language processing. ACM Transactions on Computing for Healthcare (HEALTH)  \textbf{3}(1),  1--23 (2021)

\bibitem{guan2023hallusionbench}
Guan, T., Liu, F., Li, X.W.R.X.Z., Wang, X.L.X., Yacoob, L.C.F.H.Y., Zhou, D.M.T.: Hallusionbench: An advanced diagnostic suite for entangled language hallucination \& visual illusion in large vision-language models. arXiv e-prints pp. arXiv--2310 (2023)

\bibitem{han2022wsss4luad}
Han, C., Pan, X., Yan, L., Lin, H., Li, B., Yao, S., Lv, S., Shi, Z., Mai, J., Lin, J., et~al.: Wsss4luad: Grand challenge on weakly-supervised tissue semantic segmentation for lung adenocarcinoma. arXiv preprint arXiv:2204.06455  (2022)

\bibitem{he2020pathvqa}
He, X., Zhang, Y., Mou, L., Xing, E., Xie, P.: Pathvqa: 30000+ questions for medical visual question answering. arXiv preprint arXiv:2003.10286  (2020)

\bibitem{huang2023visual}
Huang, Z., Bianchi, F., Yuksekgonul, M., Montine, T.J., Zou, J.: A visual--language foundation model for pathology image analysis using medical twitter. Nature medicine  \textbf{29}(9),  2307--2316 (2023)

\bibitem{ikezogwo2023quilt}
Ikezogwo, W.O., Seyfioglu, M.S., Ghezloo, F., Geva, D.S.C., Mohammed, F.S., Anand, P.K., Krishna, R., Shapiro, L.: Quilt-1m: One million image-text pairs for histopathology. arXiv preprint arXiv:2306.11207  (2023)

\bibitem{kather2018100}
Kather, J.N., Halama, N., Marx, A.: 100,000 histological images of human colorectal cancer and healthy tissue. Zenodo10  \textbf{5281} (2018)

\bibitem{kriegsmann2022deep}
Kriegsmann, K., Lobers, F., Zgorzelski, C., Kriegsmann, J., Janssen, C., Meliss, R.R., Muley, T., Sack, U., Steinbuss, G., Kriegsmann, M.: Deep learning for the detection of anatomical tissue structures and neoplasms of the skin on scanned histopathological tissue sections. Frontiers in Oncology  \textbf{12},  1022967 (2022)

\bibitem{kumar2014robbins}
Kumar, V., Abbas, A.K., Fausto, N., Aster, J.C.: Robbins and Cotran pathologic basis of disease, professional edition e-book. Elsevier health sciences (2014)

\bibitem{lau2018dataset}
Lau, J.J., Gayen, S., Ben~Abacha, A., Demner-Fushman, D.: A dataset of clinically generated visual questions and answers about radiology images. Scientific data  \textbf{5}(1),  1--10 (2018)

\bibitem{li2023otter}
Li, B., Zhang, Y., Chen, L., Wang, J., Yang, J., Liu, Z.: Otter: A multi-modal model with in-context instruction tuning. arXiv preprint arXiv:2305.03726  (2023)

\bibitem{li2023seed}
Li, B., Wang, R., Wang, G., Ge, Y., Ge, Y., Shan, Y.: Seed-bench: Benchmarking multimodal llms with generative comprehension. arXiv preprint arXiv:2307.16125  (2023)

\bibitem{li2023llava}
Li, C., Wong, C., Zhang, S., Usuyama, N., Liu, H., Yang, J., Naumann, T., Poon, H., Gao, J.: Llava-med: Training a large language-and-vision assistant for biomedicine in one day. arXiv preprint arXiv:2306.00890  (2023)

\bibitem{li2022yolov6}
Li, C., Li, L., Jiang, H., Weng, K., Geng, Y., Li, L., Ke, Z., Li, Q., Cheng, M., Nie, W., et~al.: Yolov6: A single-stage object detection framework for industrial applications. arXiv preprint arXiv:2209.02976  (2022)

\bibitem{li2023blip}
Li, J., Li, D., Savarese, S., Hoi, S.: Blip-2: Bootstrapping language-image pre-training with frozen image encoders and large language models. arXiv preprint arXiv:2301.12597  (2023)

\bibitem{liu2023visual}
Liu, H., Li, C., Wu, Q., Lee, Y.J.: Visual instruction tuning. arXiv preprint arXiv:2304.08485  (2023)

\bibitem{liu2023mmbench}
Liu, Y., Duan, H., Zhang, Y., Li, B., Zhang, S., Zhao, W., Yuan, Y., Wang, J., He, C., Liu, Z., et~al.: Mmbench: Is your multi-modal model an all-around player? arXiv preprint arXiv:2307.06281  (2023)

\bibitem{liu2022convnet}
Liu, Z., Mao, H., Wu, C.Y., Feichtenhofer, C., Darrell, T., Xie, S.: A convnet for the 2020s. In: Proceedings of the IEEE/CVF conference on computer vision and pattern recognition. pp. 11976--11986 (2022)

\bibitem{chatgpt}
OpenAI: Introducing chatgpt. https://openai.com/blog/chatgpt  (2022)

\bibitem{gpt4}
OpenAI: Gpt-4 technical report (2023)

\bibitem{openai2023gpt4v}
OpenAI: Gpt-4v(ision) system card. \url{https://cdn.openai.com/papers/GPTV_System_Card.pdf} (2023)

\bibitem{peng2023kosmos}
Peng, Z., Wang, W., Dong, L., Hao, Y., Huang, S., Ma, S., Wei, F.: Kosmos-2: Grounding multimodal large language models to the world. arXiv preprint arXiv:2306.14824  (2023)

\bibitem{radford2021learning}
Radford, A., Kim, J.W., Hallacy, C., Ramesh, A., Goh, G., Agarwal, S., Sastry, G., Askell, A., Mishkin, P., Clark, J., et~al.: Learning transferable visual models from natural language supervision. In: International conference on machine learning. pp. 8748--8763. PMLR (2021)

\bibitem{raffel2020exploring}
Raffel, C., Shazeer, N., Roberts, A., Lee, K., Narang, S., Matena, M., Zhou, Y., Li, W., Liu, P.J.: Exploring the limits of transfer learning with a unified text-to-text transformer. The Journal of Machine Learning Research  \textbf{21}(1),  5485--5551 (2020)

\bibitem{seyfioglu2023quilt}
Seyfioglu, M.S., Ikezogwo, W.O., Ghezloo, F., Krishna, R., Shapiro, L.: Quilt-llava: Visual instruction tuning by extracting localized narratives from open-source histopathology videos. arXiv preprint arXiv:2312.04746  (2023)

\bibitem{silva2020going}
Silva-Rodr{\'\i}guez, J., Colomer, A., Sales, M.A., Molina, R., Naranjo, V.: Going deeper through the gleason scoring scale: An automatic end-to-end system for histology prostate grading and cribriform pattern detection. Computer methods and programs in biomedicine  \textbf{195},  105637 (2020)

\bibitem{sun2021ernie}
Sun, Y., Wang, S., Feng, S., Ding, S., Pang, C., Shang, J., Liu, J., Chen, X., Zhao, Y., Lu, Y., et~al.: Ernie 3.0: Large-scale knowledge enhanced pre-training for language understanding and generation. arXiv preprint arXiv:2107.02137  (2021)

\bibitem{10095887}
Sun, Y., Zhu, C., Zhang, Y., Li, H., Chen, P., Yang, L.: Assessing the robustness of deep learning-assisted pathological image analysis under practical variables of imaging system. In: ICASSP 2023 - 2023 IEEE International Conference on Acoustics, Speech and Signal Processing (ICASSP). pp.~1--5 (2023). \doi{10.1109/ICASSP49357.2023.10095887}

\bibitem{sun2023pathasst}
Sun, Y., Zhu, C., Zheng, S., Zhang, K., Shui, Z., Yu, X., Zhao, Y., Li, H., Zhang, Y., Zhao, R., et~al.: Pathasst: Redefining pathology through generative foundation ai assistant for pathology. arXiv preprint arXiv:2305.15072  (2023)

\bibitem{team2023gemini}
Team, G., Anil, R., Borgeaud, S., Wu, Y., Alayrac, J.B., Yu, J., Soricut, R., Schalkwyk, J., Dai, A.M., Hauth, A., et~al.: Gemini: a family of highly capable multimodal models. arXiv preprint arXiv:2312.11805  (2023)

\bibitem{touvron2023llama}
Touvron, H., Lavril, T., Izacard, G., Martinet, X., Lachaux, M.A., Lacroix, T., Rozi{\`e}re, B., Goyal, N., Hambro, E., Azhar, F., et~al.: Llama: Open and efficient foundation language models. arXiv preprint arXiv:2302.13971  (2023)

\bibitem{veeling2018rotation}
Veeling, B.S., Linmans, J., Winkens, J., Cohen, T., Welling, M.: Rotation equivariant cnns for digital pathology. In: Medical Image Computing and Computer Assisted Intervention--MICCAI 2018: 21st International Conference, Granada, Spain, September 16-20, 2018, Proceedings, Part II 11. pp. 210--218. Springer (2018)

\bibitem{wang2023evaluation}
Wang, J., Zhou, Y., Xu, G., Shi, P., Zhao, C., Xu, H., Ye, Q., Yan, M., Zhang, J., Zhu, J., et~al.: Evaluation and analysis of hallucination in large vision-language models. arXiv preprint arXiv:2308.15126  (2023)

\bibitem{wang2023cogvlm}
Wang, W., Lv, Q., Yu, W., Hong, W., Qi, J., Wang, Y., Ji, J., Yang, Z., Zhao, L., Song, X., et~al.: Cogvlm: Visual expert for pretrained language models. arXiv preprint arXiv:2311.03079  (2023)

\bibitem{wei2021petri}
Wei, J., Suriawinata, A., Ren, B., Liu, X., Lisovsky, M., Vaickus, L., Brown, C., Baker, M., Tomita, N., Torresani, L., et~al.: A petri dish for histopathology image analysis. In: Artificial Intelligence in Medicine: 19th International Conference on Artificial Intelligence in Medicine, AIME 2021, Virtual Event, June 15--18, 2021, Proceedings. pp. 11--24. Springer (2021)

\bibitem{xu2023lvlm}
Xu, P., Shao, W., Zhang, K., Gao, P., Liu, S., Lei, M., Meng, F., Huang, S., Qiao, Y., Luo, P.: Lvlm-ehub: A comprehensive evaluation benchmark for large vision-language models. arXiv preprint arXiv:2306.09265  (2023)

\bibitem{yin2023lamm}
Yin, Z., Wang, J., Cao, J., Shi, Z., Liu, D., Li, M., Sheng, L., Bai, L., Huang, X., Wang, Z., et~al.: Lamm: Language-assisted multi-modal instruction-tuning dataset, framework, and benchmark. arXiv preprint arXiv:2306.06687  (2023)

\bibitem{yu2023mm}
Yu, W., Yang, Z., Li, L., Wang, J., Lin, K., Liu, Z., Wang, X., Wang, L.: Mm-vet: Evaluating large multimodal models for integrated capabilities. arXiv preprint arXiv:2308.02490  (2023)

\bibitem{yue2023mmmu}
Yue, X., Ni, Y., Zhang, K., Zheng, T., Liu, R., Zhang, G., Stevens, S., Jiang, D., Ren, W., Sun, Y., et~al.: Mmmu: A massive multi-discipline multimodal understanding and reasoning benchmark for expert agi. arXiv preprint arXiv:2311.16502  (2023)

\bibitem{zhang2023pmc}
Zhang, X., Wu, C., Zhao, Z., Lin, W., Zhang, Y., Wang, Y., Xie, W.: Pmc-vqa: Visual instruction tuning for medical visual question answering. arXiv preprint arXiv:2305.10415  (2023)

\bibitem{zhang2022benchmarking}
Zhang, Y., Sun, Y., Li, H., Zheng, S., Zhu, C., Yang, L.: Benchmarking the robustness of deep neural networks to common corruptions in digital pathology. In: International Conference on Medical Image Computing and Computer-Assisted Intervention. pp. 242--252. Springer (2022)

\bibitem{zhang2019pathologist}
Zhang, Z., Chen, P., McGough, M., Xing, F., Wang, C., Bui, M., Xie, Y., Sapkota, M., Cui, L., Dhillon, J., et~al.: Pathologist-level interpretable whole-slide cancer diagnosis with deep learning. Nature Machine Intelligence  \textbf{1}(5),  236--245 (2019)

\bibitem{zheng2024benchmarking}
Zheng, S., Cui, X., Sun, Y., Li, J., Li, H., Zhang, Y., Chen, P., Jing, X., Ye, Z., Yang, L.: Benchmarking pathclip for pathology image analysis (2024)

\bibitem{zhu2022weakly}
Zhu, C., Sun, Y., Li, H., Cui, C., Zhang, S., Cai, J., Ling, Y.: Weakly supervised classification using multi-level instance-aware optimization on cervical cytologic image. In: 2022 IEEE 19th International Symposium on Biomedical Imaging (ISBI). pp.~1--5. IEEE (2022)

\bibitem{zhu2023minigpt}
Zhu, D., Chen, J., Shen, X., Li, X., Elhoseiny, M.: Minigpt-4: Enhancing vision-language understanding with advanced large language models. arXiv preprint arXiv:2304.10592  (2023)

\end{thebibliography}
\clearpage
\appendix
\section{Supplementary Information of PathMMU}

\subsection{Statistical Distribution}
In this section, we present an overview of the fundamental statistics for the PathMMU dataset. This includes the total count of questions and images, the average length of questions, the average length of options, the average resolution of images, and a detailed analysis of word frequency.

\textbf{Basic Statistical Information}: 
\cref{tab: PathMMU train set distribution} and \cref{tab: PathMMU test set distribution} showcase the statistical information for the PathMMU train and test sets. Key observations include: \textit{\textbf{High Diversity and Quantity}}: Since each image encompasses a limited scope of knowledge points, associating multiple Q\&As with a single image could result in a narrow and similar range of examined knowledge areas. Therefore, one effective strategy to enhance the dataset's diversity is to increase the number and sources of images. In essence, given a fixed number of questions, a larger selection of images and sources significantly broadens the overall spectrum of knowledge covered. Therefore, PathMMU's exceptionally low question-to-image ratio, and diverse sources, coupled with the substantial volume of QA pairs, highlight its strengths in terms of both diversity and quantity.
\textit{\textbf{Exceptional Clarity}}: 
The images in our collection stand out for their exceptionally high resolution and are primarily composed of single figures, instead of sub-figures. Each image undergoes a rigorous manual review for clarity, ensuring the quality and distinctiveness of every visual element within our dataset. Among them, the images sourced from YouTube educational content typically boast higher resolutions, largely owing to them being screenshots taken from screen-sharing sessions employed in educational materials for pathologists.

\textbf{Word Frequency Visualization}: 
\cref{fig:twitter_cap} illustrates the word frequency in the original captions of image pairs from the SocialPath subset, while \cref{fig:twitter_des} demonstrates the word frequency in image descriptions. Notably, post-processing and caption refinement with GPT-4V led to a significant decrease in irrelevant terms such as `Pathtwitter' and `Gipath', shifting the focus more towards pathology-specific terminology like `cell' and `tissue.' This change signifies a more targeted emphasis on pathology-related features in the processed data. \cref{fig:test_all_question} visualizes the word frequencies in the questions. Certain key terms, including `what', `which', `observed', `cells' and `tissue', stand out prominently, highlighting a significant emphasis on questions related to specific tissue characteristics and diagnostic procedures. Furthermore, \cref{fig:test_all_option} illustrates the word frequencies in the answer options, revealing a noticeable focus on morphological features and diagnostic terms. This pattern indicates that the questions and options are designed to deeply explore the valuable aspects of pathology and diagnosis, thereby exemplifying the professionalism and diversity of our PathMMU dataset.

\begin{table}[]
	\centering
        \caption{Statistics of each subset in the PathMMU train set.}
	\resizebox{0.85\columnwidth}{!}{%
		\begin{tabular}{@{}ccccc@{}}
			\toprule
			Source       & \textbf{\begin{tabular}[c]{@{}c@{}}Avg. Question \\ Length\end{tabular}} & \textbf{\begin{tabular}[c]{@{}c@{}}Avg. Option \\ Length\end{tabular}} & \textbf{\begin{tabular}[c]{@{}c@{}}\# Questions \\ / \# Images\end{tabular}} & \textbf{\begin{tabular}[c]{@{}c@{}} Avg. Image\\ Resolution\end{tabular}} \\ \midrule
                PubMed & 16.7 & 4.5 & 6064 / 4342     & 702 $\times$ 558      \\
			SocialPath      & 16.2 & 4.7 & 7381 / 4606     & 1084 $\times$ 861     \\
			EduContent      & 16.1 & 4.7 & 5214 / 3250     & 1573 $\times$ 888     \\
			Atlas         & 16.5 & 4.2 & 622 / 386     & 589 $\times$ 451      \\           
			PathCLS & 20.1 & 3.2 & 3760 / 3760     & 526 $\times$ 494      \\

			Overall      & 17.0 & 4.4 & 23041 / 16344   & 940 $\times$ 692     \\ \bottomrule
		\end{tabular}
	}
	\label{tab: PathMMU train set distribution}
\end{table}

\begin{table}[]
	\centering
        \caption{Statistics of each subset in the PathMMU test set.}
	\resizebox{0.85\columnwidth}{!}{%
		\begin{tabular}{@{}ccccc@{}}
			\toprule
			Source       & \textbf{\begin{tabular}[c]{@{}c@{}}Avg. Question \\ Length\end{tabular}} & \textbf{\begin{tabular}[c]{@{}c@{}}Avg. Option \\ Length\end{tabular}} & \textbf{\begin{tabular}[c]{@{}c@{}}\# Questions \\ / \# Images\end{tabular}} & \textbf{\begin{tabular}[c]{@{}c@{}} Avg. Image\\ Resolution\end{tabular}} \\ \midrule
                PubMed & 16.5 & 4.4 & 3068 / 2098 & 704$\times$558 \\
			SocialPath      & 16.3 & 4.3 & 1855 / 1218 & 1074$\times$852 \\
			EduContent      & 16.2 & 4.5 & 1938 / 1447 & 1451$\times$844 \\
			Atlas         & 16.3 & 4.3 & 1007 / 641  & 585$\times$465  \\           
			
			PathCLS& 20.4 & 3.1 & 1809 / 1809 & 560$\times$517 \\
			Overall      & 17.1 & 4.2 & 9677 / 7213 & 874$\times$650 \\ \bottomrule
		\end{tabular}%
	}
	\label{tab: PathMMU test set distribution}
\end{table}

\subsection{Overview of the Composition of the PathCLS Subset}

As detailed in \cref{sec:pathmmu_construction}, the PathCLS subset is developed by gathering data from ten distinct pathology classification datasets. Below, we provide a comprehensive overview of each dataset: \textbf{CRC100K}~\cite{kather2018100}: Comprising image patches extracted from H\&E stained histological samples, this dataset encompasses both colorectal cancer tissues and normal tissues. It is organized into nine categories: Adipose, Background, Debris, Lymphocytes, Mucus, Smooth Muscle, Normal Colon Mucosa, Cancer-Associated Stroma, and Colorectal Adenocarcinoma Epithelium. \textbf{WSSS4LUAD}~\cite{han2022wsss4luad}: This collection includes patch-level annotations from 87 whole slide images, focusing on differentiating between tumor and normal tissue classes. \textbf{LC25000}~\cite{borkowski2019lung}: Featuring samples of lung and colon adenocarcinomas, the dataset is divided into two subsets: \textbf{LC-lung and LC-colon}. The LC-lung section includes lung adenocarcinomas, lung squamous cell carcinomas, and benign lung tissues. The LC-colon segment comprises colon adenocarcinomas and benign colonic tissues. \textbf{PatchCamelyon}~\cite{veeling2018rotation}: Derived from histopathological scans of lymph node sections, each image in this dataset is marked with a binary label to indicate the presence or absence of metastatic tissue. \textbf{SICAPv2}~\cite{silva2020going}: This dataset contains images of prostate pathology magnified 10 times, classified into non-cancerous and Grades 3-5 according to the Gleason grading system. \textbf{BACH}~\cite{aresta2019bach}: Incorporating H\&E stained breast histology images, this dataset categorizes images into normal, benign, in situ carcinoma, or invasive carcinoma based on the dominant cancer type present in each image. \textbf{Osteo}~\cite{arunachalam2019viable}: Sourced from whole slide images that capture the diverse nature of osteosarcoma and its response to chemotherapy, this dataset aims to classify different tissue regions as viable tumors, necrotic tumors, or non-tumors. \textbf{SkinCancer}~\cite{kriegsmann2022deep}: This dataset consists of tissue patches from skin biopsies spanning 12 anatomical compartments and 4 types of neoplasms. \textbf{MHIST}~\cite{wei2021petri}: Featuring tissue patches from Formalin-Fixed Paraffin-Embedded whole slide images of colorectal polyps.

\begin{figure*}[!t]
	\centering
	\includegraphics[width=\linewidth]{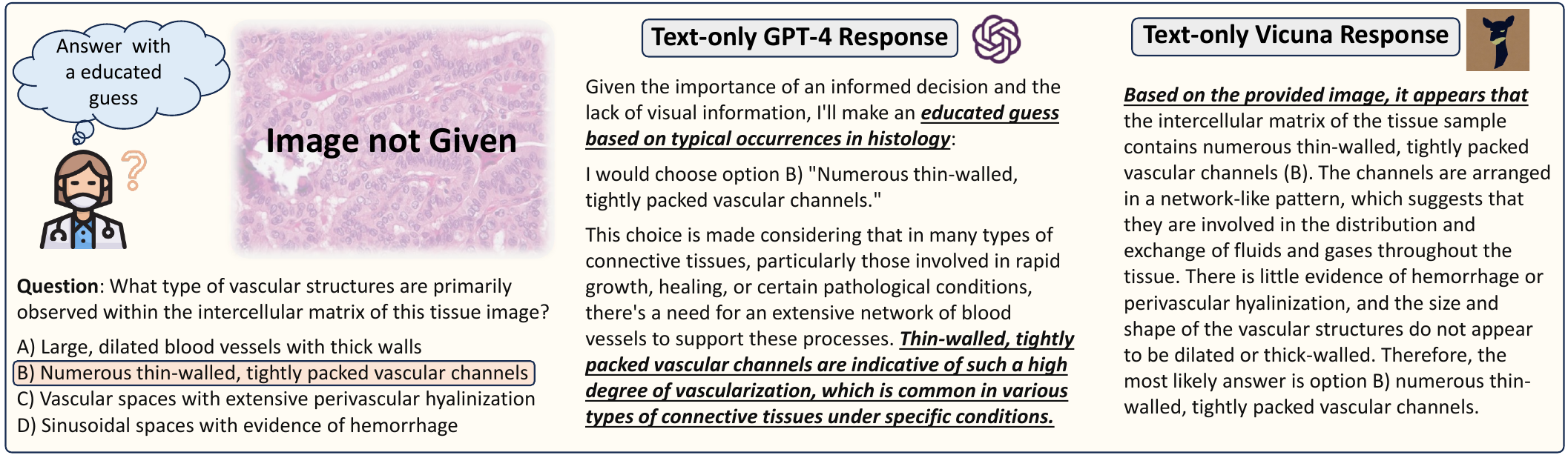}
	\caption{Examples of how GPT-4 and Vicuna respond to questions without access to images. GPT-4 admits that its answers are based on educated guesses, while Vicuna experiences hallucinations, believing that it has visual access to the images.}
	\label{fig: guess illustration}
\end{figure*}

\section{Supplementary Experimental Findings}
\subsection{Comparison of GPT-4 and Vicuna in Performing Text-only Educated Guess}
To compare the logic behind the most advanced closed-source LLM and the most representative open-source Vicuna-v1.5-13B model in performing educated guesses, we removed the images provided by PathMMU in the multimodal Q\&A, allowing them to make guesses based solely on text. As shown in \cref{fig: guess illustration}, GPT-4, when making guesses, acknowledges that its guesses are based on `typical occurrences in histology'. This contrasts with the approach of the smaller open-source model, Vicuna-v1.5-13B, which generates hallucinations and falsely claims visual access to the provided image.  Interestingly, Vicuna behaves as though it has a visual perception of the images, making guesses and fabricating observations to support its answers. We hypothesize that the responses of both GPT-4 and Vicuna are influenced by the commonality of certain options or their frequency in their respective training datasets. The key distinction is that GPT-4 acknowledges its inability to visually interpret images and relies on educated guesses, whereas Vicuna undergoes hallucinations, responding under the false belief of having visual access to the images.

\begin{figure*}[t!]
	\centering
	\includegraphics[width=\linewidth]{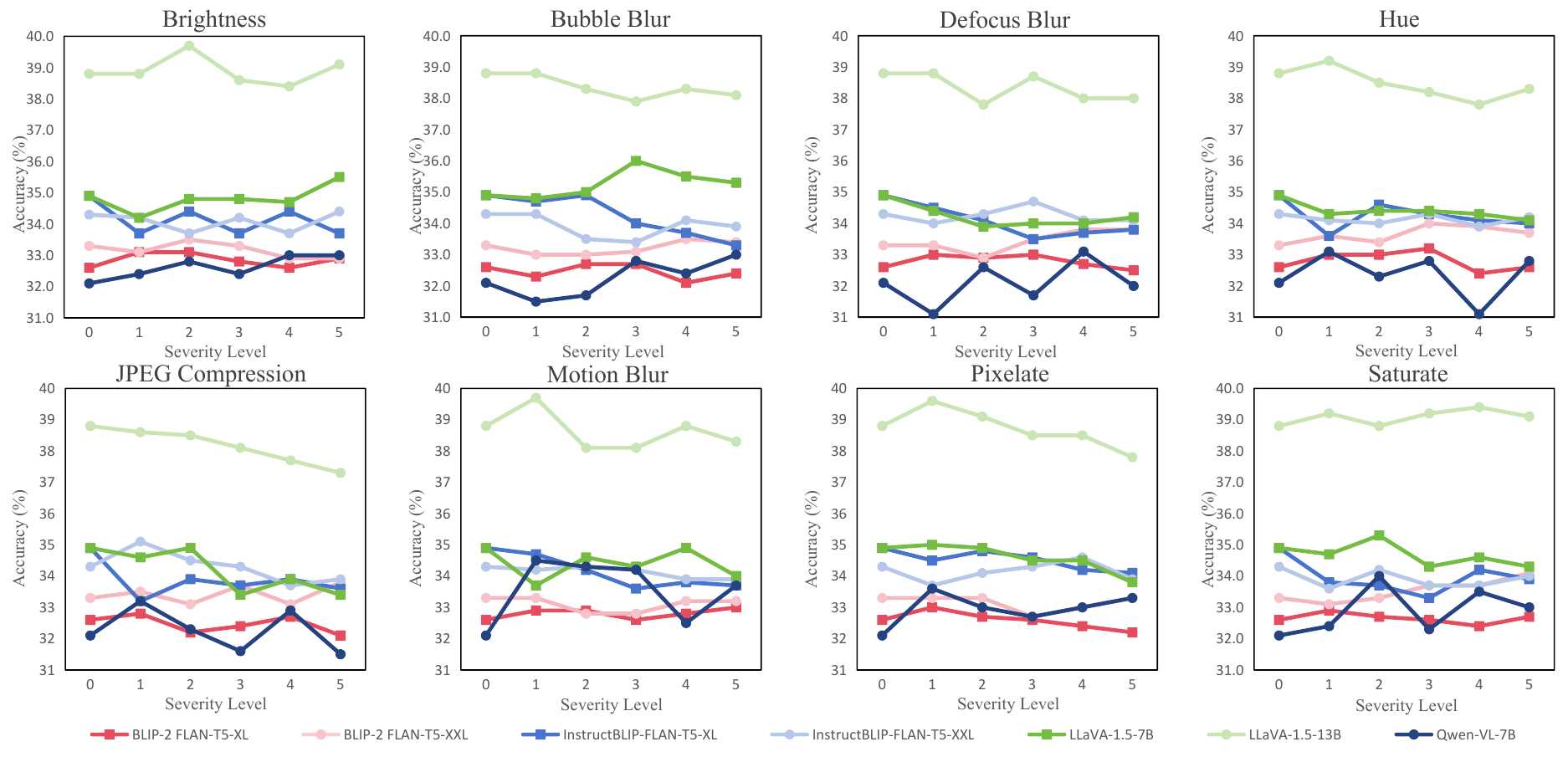}
	\caption{Illustration of the model’s performance towards different levels of corruption on the PathMMU test-tiny set, where level 0 represents the model’s performance without any image corruption.}
	\label{fig:performance_towards_corruption_level}
\end{figure*}

\subsection{LMMs' Performance Degradation towards Different Levels of Image Corruption}
\cref{fig:corruption_level} illustrates visualizations of various corruptions across a range of image corruption severity levels. \cref{fig:performance_towards_corruption_level} demonstrates the performance fluctuations of different models as the severity of corruption changes. It is noticeable that from corruption level 0 to 5, there is no significant downward trend in performance, suggesting that these corruptions have a minimal impact on the performance of LMMs on PathMMU, which supports the hypothesis proposed in \cref{sec:robustness}.

\section{Case Study}
Please kindly refer to the \textbf{project page}, we show the detailed image description generation cases, invalid cases, GPT-4V error cases and correct cases.

\section{Limitations}
In the field of pathology, diagnosing pathological images is often influenced by the following factors: \textit{\textbf{(1) Individual Differences:}} Different pathologists have variations in their understanding and application of diagnostic criteria. Even among experienced pathologists, interpretations of specific criteria vary. For instance, in liquid-based cervical cytology, whether a nucleus enlargement of 2.5-3 times or more than 3 times influences the diagnosis of Atypical Squamous Cells of Undetermined Significance (ASC-US) or Low-grade Squamous Intraepithelial Lesion (LSIL) to some extent. However, the understanding of the value 2.5 differs among pathologists, ultimately leading to different diagnoses. \textit{\textbf{(2) Emotional and Physical State:}} A pathologist's judgment can also be affected by their emotional state, level of fatigue, and other physical conditions, which may lead to variations in the reading of the same sample at different times. \textit{\textbf{(3) Specialty Preferences:}} Pathologists may have more experience and expertise in certain subspecialties of pathology (such as breast pathology, neuropathology, \etc), which can affect their accuracy in interpreting samples from other areas.

\begin{table}[t]
	\addtolength{\tabcolsep}{-0.9pt}
	\centering
	\caption{The validity assessments of the same 100 samples by two pathologists.}
	\resizebox{0.5\columnwidth}{!}{%
		\begin{tabular}{@{}cccc@{}}
			\toprule
			& \textbf{Valid} & \textbf{Invalid} & \textbf{Consistency}  \\ \midrule
			Pathologist A & 83             & 17               & \multirow{2}{*}{90\%} \\
			Pathologist B & 89             & 11               &                       \\ \bottomrule
		\end{tabular}%
	}
\label{tab:consistency of pathologists}
\end{table}

These factors impact the construction of our dataset in two significant ways: \textit{\textbf{(1) }}During the process of selecting effective data, these influencing factors can lead to the excessive elimination of some valid samples or the omission of some invalid ones. To address this, we conduct cross-experiments where 100 images are annotated by two pathologists. As shown in \cref{tab:consistency of pathologists}, the agreement rate between the two pathologists is 90\%, indicating that there are invalid samples within the samples deemed valid by pathologist A, and vice versa. \textit{\textbf{(2) }}The benchmark of human experts is not the upper limit of the dataset. Because a single doctor cannot be highly proficient in all disciplines of pathology, we select some of the most comprehensive doctors to conduct this test, even though they cannot cover the entirety of the field's expertise.

\newpage
\begin{figure*}[t]
	\centering
	\includegraphics[width=\linewidth]{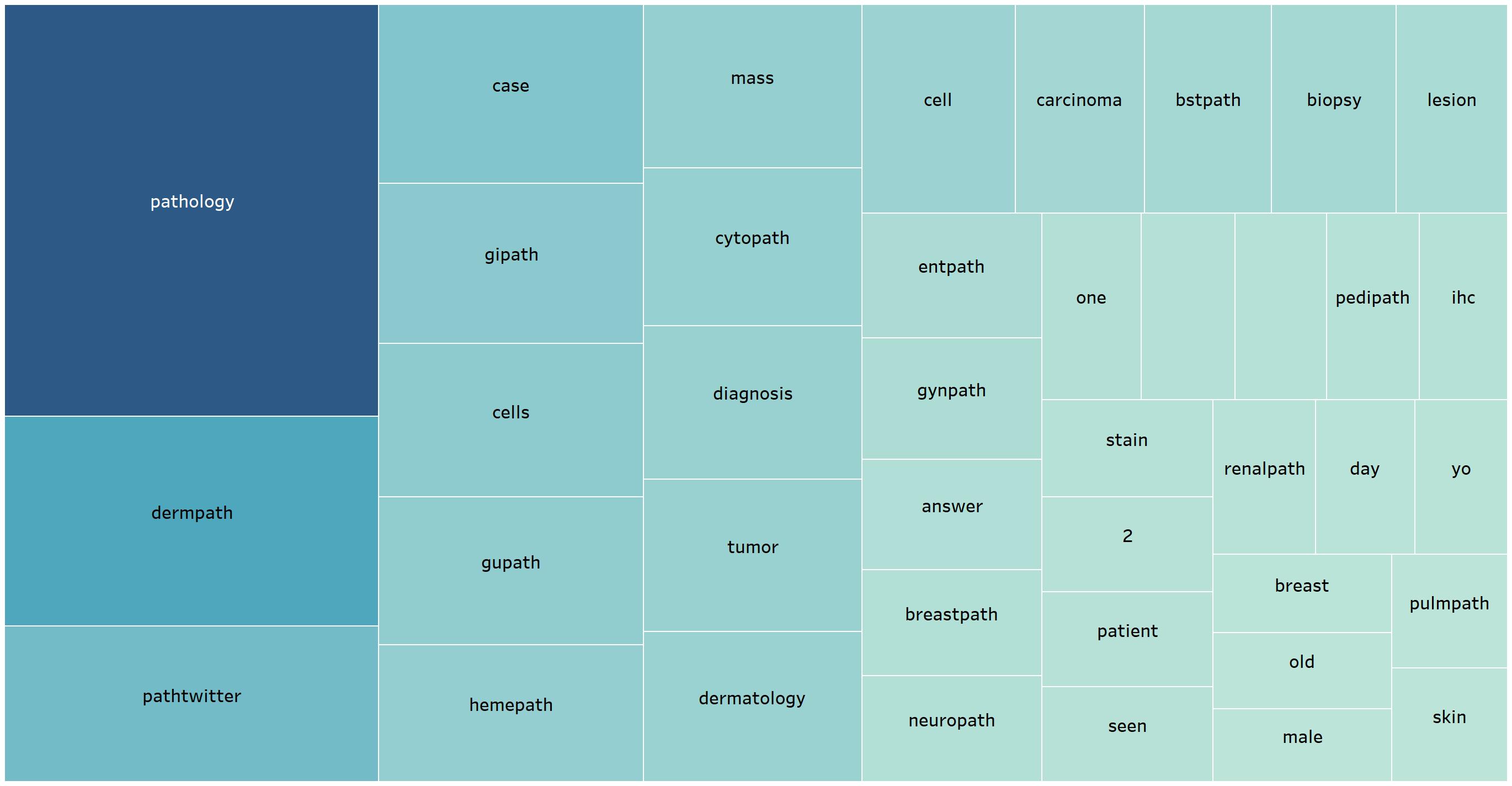}
	\caption{Visualization of the top 40 word frequencies in captions from SocialPath subset.}
	\label{fig:twitter_cap}
\end{figure*}
\begin{figure*}[h]
	\centering
	\includegraphics[width=\linewidth]{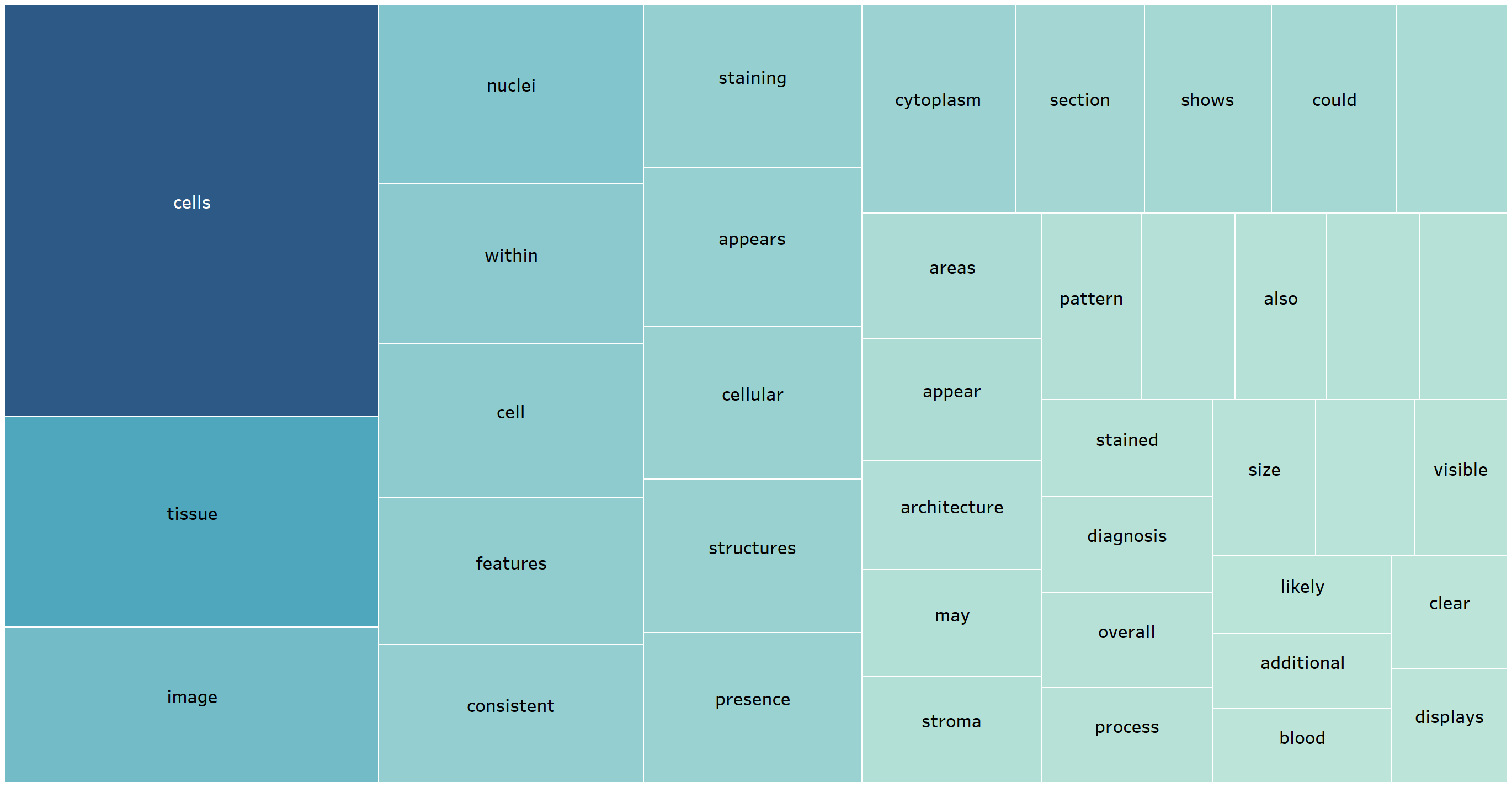}
	\caption{Visualization of the top 40 word frequencies in generated descriptions from SocialPath subset.}
	\label{fig:twitter_des}
\end{figure*}

\begin{figure*}[]
    \centering
    \includegraphics[width=\linewidth]{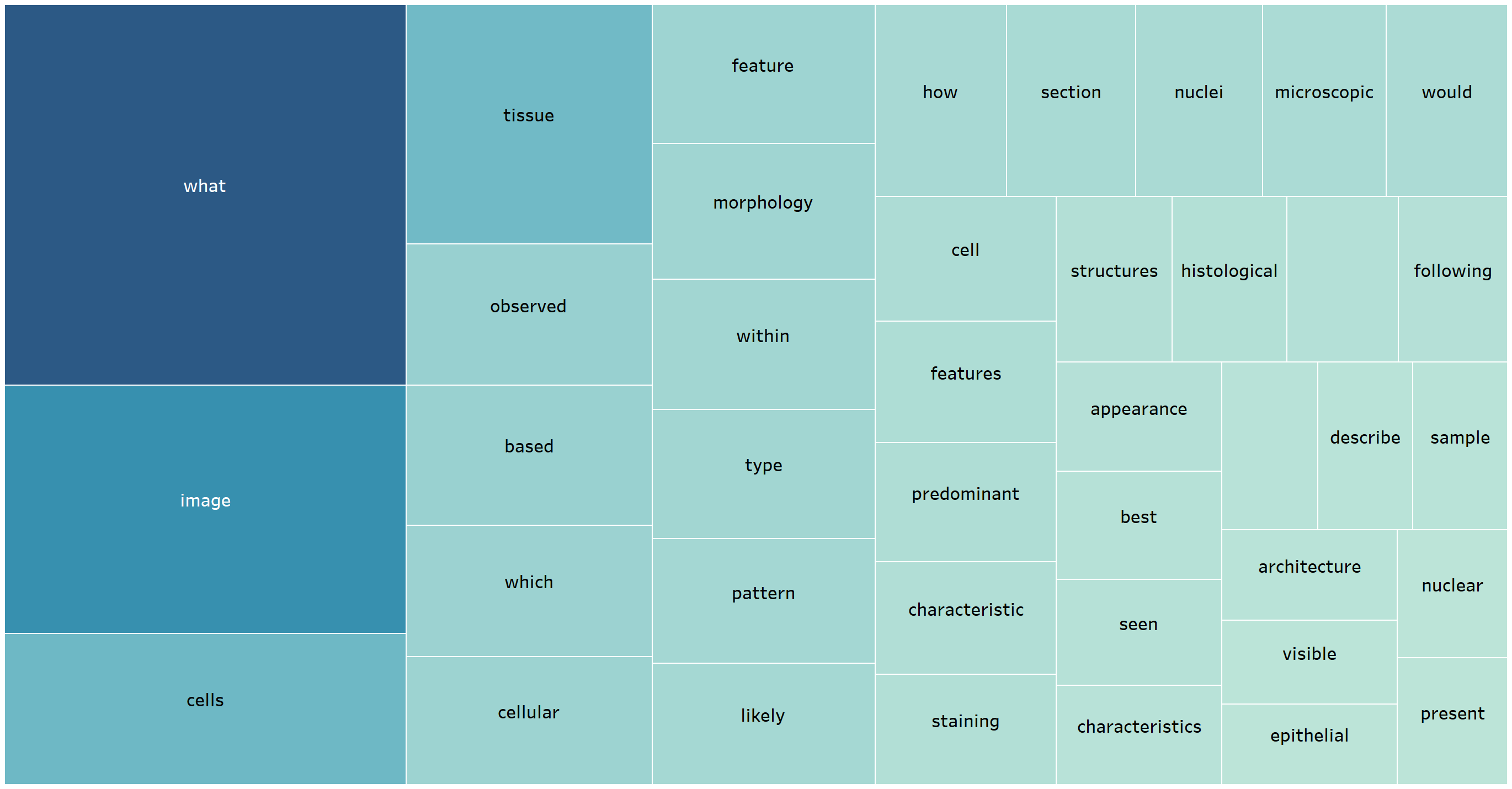}
    \caption{Visualization of the top 40 most frequent words in questions from the PathMMU test set.}
    \label{fig:test_all_question}
\end{figure*}

\begin{figure*}[]
    \centering
    \includegraphics[width=\linewidth]{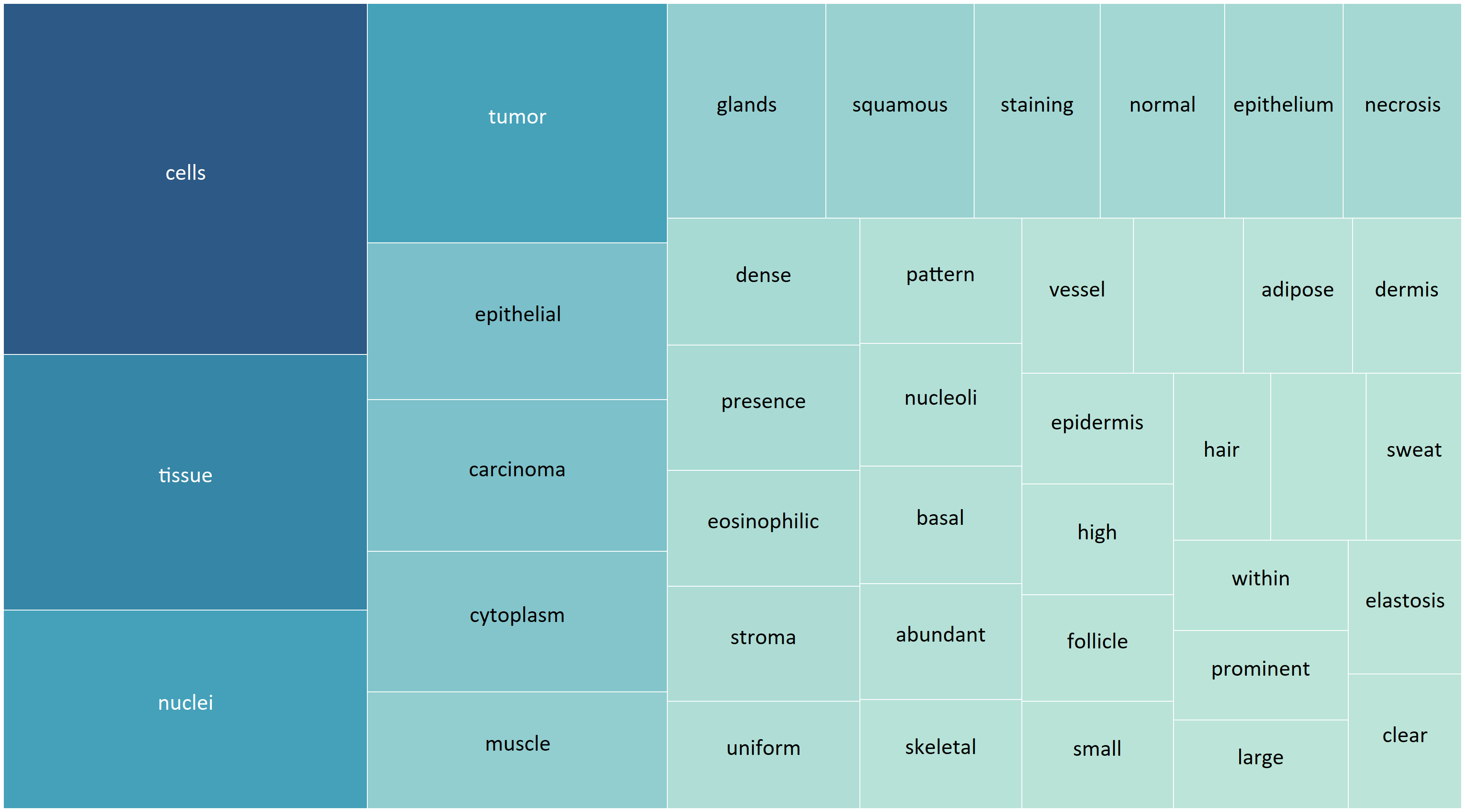}
    \caption{Visualization of the top 40 most frequent words in options from the PathMMU test set.}
    \label{fig:test_all_option}
\end{figure*}

\begin{figure*}[!t]
	\centering
	\includegraphics[width=\linewidth]{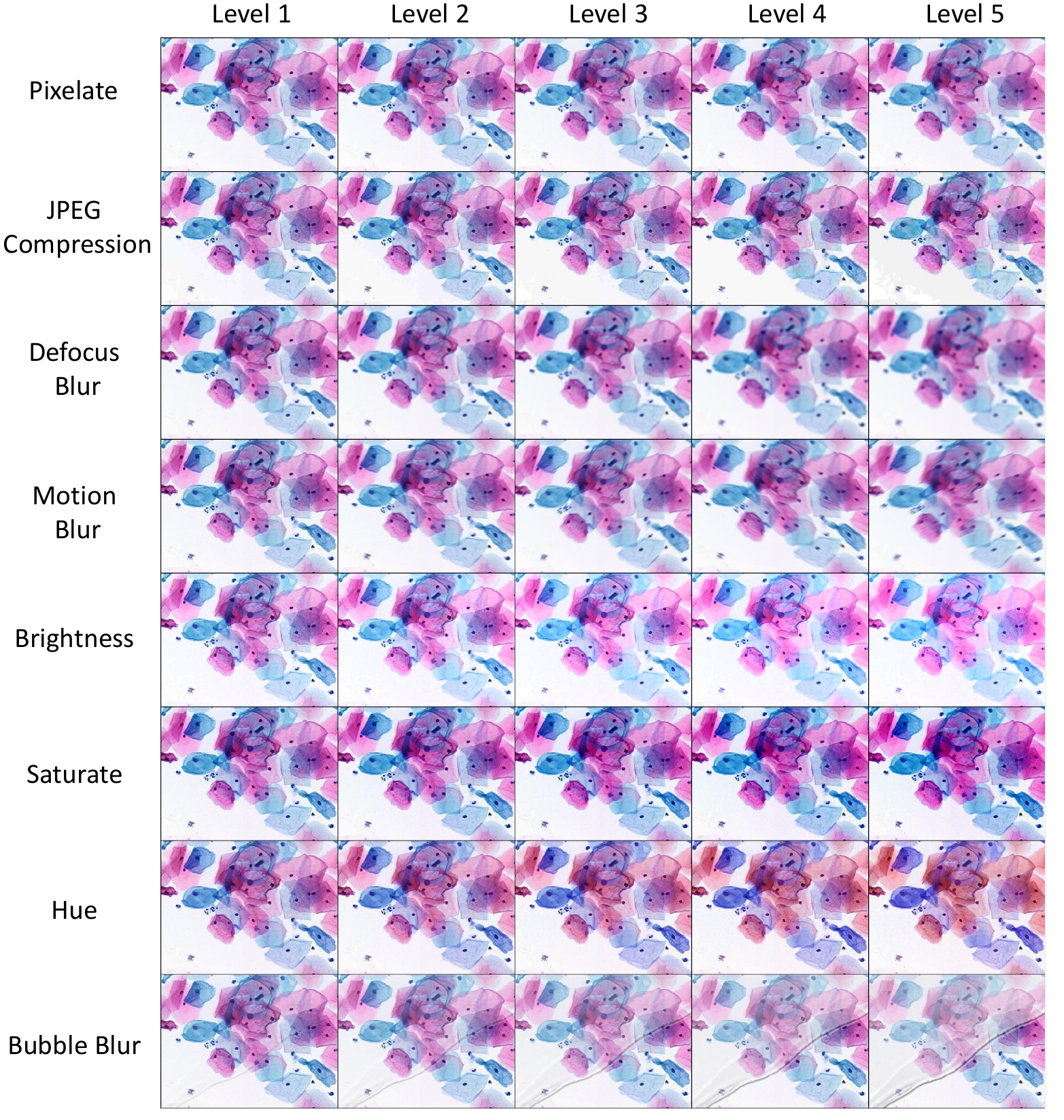}
	\caption{Examples of pathology images with five different levels of corruption severity.}
	\label{fig:corruption_level}
\end{figure*}

\end{document}